\pdfoutput=1

\documentclass{article}

\usepackage[preprint]{neurips_2023}
\usepackage{natbib}

\usepackage{times}
\usepackage{latexsym}

\usepackage[T1]{fontenc}

\usepackage[utf8]{inputenc}

\usepackage{microtype}

\usepackage{inconsolata}

\usepackage{amsmath}
\usepackage{algorithm}
\usepackage{algpseudocode}
\usepackage{url}
\usepackage{booktabs}
\usepackage{multicol}
\usepackage{multirow}
\usepackage{color}
\usepackage{xcolor}     
\usepackage{colortbl}
\definecolor{mydarkblue}{rgb}{0,0.08,0.45}
\usepackage[colorlinks=true,linkcolor=mydarkblue,citecolor=mydarkblue,filecolor=mydarkblue,urlcolor=mydarkblue]{hyperref}
\usepackage{cleveref}
\usepackage{todonotes}
\usepackage{csquotes}
\usepackage{arabtex}
\usepackage{utf8}
\setcode{utf8}
\renewcommand{\cite}{\citep}
\usepackage{pgfplots}
\usepackage{pgf}
\usepackage{lmodern}
\pgfplotsset{compat=1.16}

\newcommand{\best}[1]{\textbf{#1}}

\title{ClimateGPT: Towards AI Synthesizing Interdisciplinary Research on Climate Change}

\author{
\textbf{David Thulke}\textsuperscript{\rm 1,4} \quad
\textbf{Yingbo Gao}\textsuperscript{\rm 1} \quad
\textbf{Petrus Pelser}\textsuperscript{\rm 3} \quad
\textbf{Rein Brune}\textsuperscript{\rm 3}  \\
\textbf{Rricha Jalota}\textsuperscript{\rm 1} \quad
\textbf{Floris Fok}\textsuperscript{\rm 3} \quad
\textbf{Michael Ramos}\textsuperscript{\rm 2}  \quad
\textbf{Ian van Wyk}\textsuperscript{\rm 3} \\
\textbf{Abdallah Nasir}\textsuperscript{\rm 1} \quad
\textbf{Hayden Goldstein}\textsuperscript{\rm 3} \quad
\textbf{Taylor Tragemann}\textsuperscript{\rm 1}  \quad
\textbf{Katie Nguyen}\textsuperscript{\rm 1} \\
\textbf{Ariana Fowler}\textsuperscript{\rm 2} \quad
\textbf{Andrew Stanco}\textsuperscript{\rm 2} \quad
\textbf{Jon Gabriel}\textsuperscript{\rm 2}  \quad
\textbf{Jordan Taylor}\textsuperscript{\rm 2} \\
\textbf{Dean Moro}\textsuperscript{\rm 2} \quad
\textbf{Evgenii Tsymbalov}\textsuperscript{\rm 1} \quad
\textbf{Juliette de Waal}\textsuperscript{\rm 3} \quad
\textbf{Evgeny Matusov}\textsuperscript{\rm 1} \\
\textbf{Mudar Yaghi}\textsuperscript{\rm 1} \quad
\textbf{Mohammad Shihadah}\textsuperscript{\rm 1} \quad
\textbf{Hermann Ney}\textsuperscript{\rm 1,4} \\
\textbf{Christian Dugast}\textsuperscript{\rm 1} \quad
\textbf{Jonathan Dotan}\textsuperscript{\rm 2} \quad
\textbf{Daniel Erasmus}\textsuperscript{\rm 3}\vspace{1mm}\\
\textsuperscript{\rm 1}AppTek \quad
\textsuperscript{\rm 2}EQTY Lab \quad
\textsuperscript{\rm 3}Erasmus.AI \quad
\textsuperscript{\rm 4}RWTH Aachen University \\
\texttt{dthulke@apptek.com} \quad \texttt{climategpt@dtn.net} \\
  \texttt{\url{eci.io}}
}

\begin{document}

\maketitle

\begin{abstract}
This paper introduces ClimateGPT, a model family of domain-specific large language models that synthesize interdisciplinary research on climate change.
We trained two 7B models from scratch on a science-oriented dataset of 300B tokens.
For the first model, the 4.2B domain-specific tokens were included during pre-training and the second was adapted to the climate domain after pre-training.
Additionally, ClimateGPT-7B, 13B and 70B are continuously pre-trained from Llama~2 on a domain-specific dataset of 4.2B tokens.
Each model is instruction fine-tuned on a high-quality and human-generated domain-specific dataset that has been created in close cooperation with climate scientists.
To reduce the number of hallucinations, we optimize the model for retrieval augmentation and propose a hierarchical retrieval strategy.
To increase the accessibility of our model to non-English speakers, we propose to make use of cascaded machine translation and show that this approach can perform comparably to natively multilingual models while being easier to scale to a large number of languages.
Further, to address the intrinsic interdisciplinary aspect of climate change we consider different research perspectives.
Therefore, the model can produce in-depth answers focusing on different perspectives in addition to an overall answer.
We propose a suite of automatic climate-specific benchmarks to evaluate LLMs.
On these benchmarks, ClimateGPT-7B performs on par with the ten times larger Llama-2-70B Chat model while not degrading results on general domain benchmarks.
Our human evaluation confirms the trends we saw in our benchmarks.
All models were trained and evaluated using renewable energy and are released publicly\footnote{\url{https://huggingface.co/eci-io/}}.
\end{quote} %
\end{abstract}

\newpage

\tableofcontents

\newpage

\begin{figure}
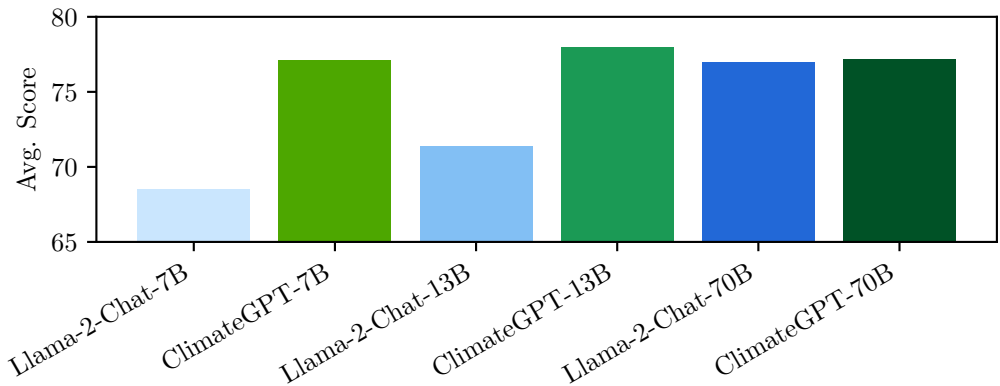

    \centering
    \include{figures/overview_bar_plot.pgf}
    \vspace{-2em}
    \caption{Overview of automatic evaluation results on climate-specific benchmarks.}
    \label{fig:overview_automatic_eval}
\end{figure}

\section{Introduction}

Large Language Models (LLMs) have the exceptional ability to comprehend and generate human-like text that empowers them to address a wide array of tasks with Claude-2\footnote{\url{https://www.anthropic.com/index/claude-2}}, GPT-4 \cite{openai2023gpt4}, Llama-2 \cite{touvron2023llama2} or Gemini \cite{geminiteam2023gemini} to cite a few.
They have been trained on diverse large datasets (from hundreds of billions to trillions of tokens) covering a wide range of topics and domains.
The universality of these general-purpose models has made them accessible for a broad spectrum of applications: from text comprehension, over content generation and summarization up to chatbots and much more. 
Recent research has pointed to the potential of LLMs trained on domain-specific data, e.g. Biomedical sciences \cite{Lee2020biobert}, Finance \cite{Wu2023bloombergGPT} and Medicine \cite{Peng2023astudy, Luo2022BioGPT}.
These models, while being smaller, have outperformed general-purpose models in their respective domains.
The work reported in this paper continues this line of research, addressing one of the most pressing and complex challenges this time: climate change.

Climate change stands out as a multifaceted discipline, covering climate science (the natural science behind modeling climate and the development of the earth's atmosphere) as well as human issues related to climate that impact our environment, our economies, our societies, public health and biodiversity.
Right now, we are moving to the brink of multiple risk tipping points \cite{UNU-EHS2023risk}.
Efforts are underway to avoid getting at these potentially irreversible changes in the climate system.
Accelerating this process requires global climate awareness and collective knowledge, that we call \enquote{climate social intelligence}.
Building an LLM that addresses climate questions requires access to this collective knowledge, understanding, and decision-making capacity of the human population to harness the collective climate social intelligence available.   

We propose an LLM on climate change, called ClimateGPT, which should help the diverse science communities involved to exchange information and knowledge along the three major multi-disciplinary dimensions it covers at large: environmental and natural science, economics, and social science.
As opposed to other work done around climate-related LLMs, e.g. ClimateBERT \cite{webersinke2022climatebert}, ClimateGPT-2 \cite{vaghefi2022deep}, MBZUAI Arabic Mini-ClimateGPT \cite{Mullappilly2023arabic}, ChatClimate \cite{chatclimate}, the focus of our work was to develop high quality in-domain Instruction Fine-Tuning (IFT) data as well as to train our model with as much climate data as possible, specifically technical reports from the Intergovernmental Panel on Climate Change (IPCC) as well as top papers from climate change research and related fields, such as the UN Sustainable Development Goals.
Further, we developed a multi-domain Large Language Model, which can give four types of answers for each request: a natural science-related answer, an answer about the economic aspects of climate change, as well as an answer about social impacts. The fourth answer, the main one, gives a general high-level overview, addressing all of these sub-fields. 

This paper introduces a large language model that seeks to be used across domains by people learning from and collaborating with other specialists in the realm of climate information, rather than merely acting as a chatbot. We are looking at it as a climate intelligence platform that can assist governments, organizations, and individuals in making informed decisions and that contributes to a global social intelligence related to climate.

\subsection{Technical Approach}

This section outlines our technical approach and the different steps we have taken to develop ClimateGPT.

\textbf{Language Modeling} is done with a large decoder-only Transformer \citep{vaswani2017attention,j.2018generating} architecture, which is in line with most of the recent literature on large language models (LLMs) \cite{radford2019language, brown2020language, workshop2022bloom, biderman2023pythia}.
The model represents tokens as continuous-valued hidden vectors and makes use of the attention mechanism \cite{bahdanau2014neural} to model inter-token dependencies.
The training criterion is cross-entropy, which rewards the model for high probability mass on the correct target token.

\textbf{From-Scratch (FS)} training is done to obtain a foundation language model in the climate domain, the training data for which is cleaned with the climate domain in mind.
We train a climate foundation model as well as a general domain model with a focus on scientific content to study the effect of up-sampling domain-specific data during foundation model training.
We follow closely the training hyper-parameters that were documented in the Llama-2 paper \cite{touvron2023llama2}.

\textbf{Continued Pre-Training (CPT)} is a common alternative to training a new foundation model from scratch \cite{gupta2023continual, chen2023meditron70b}.
The goal is to adapt an existing LLM trained on a large set of general domain data to the target domain by continuing the pre-training on a smaller set of in-domain data.
After an initial evaluation, we focus on the Llama-2 model series as well as for our general not climate-specific from-scratch model.
During CPT, we keep the training criterion of the pre-trained model.

\textbf{Instruction Fine-Tuning (IFT)} is an important step to inject instruction-following capabilities into the model.
In the literature, this is also often referred to as Supervised Fine-Tuning (SFT).
We prefer IFT to make the distinction to domain adaptation via CPT or task-specific supervised fine-tuning approaches.
Instruction-completion pairs both from the general domain and climate domain are prepared and gear the model towards following user instructions.
We collaborate with climate experts to create a high-quality human-generated dataset.
During the data collection, we follow standard approaches \cite{ouyang2022training} and also tune the distribution among our and different public instruction-tuning datasets.
Although our model is capable of chatting, we focus on its question-answering and instruction-following aspects, which also greatly simplify the instruction fine-tuning data creation and retrieval steps.

\textbf{Retrieval Augmented Generation (RAG)} is implemented with high-quality climate resources to increase factuality as well as to extend the system with new knowledge.
We crawl text from manually curated sources and process these sources into smaller chunks.
To retrieve relevant documents for a user query, we use a bi-encoder model to calculate embeddings and make use of efficient nearest-neighbor search.
During the generation phase, the user instruction is concatenated with the most relevant text chunks for the model to come up with more reliable and stable answers.
As the sources of retrieved documents are known, RAG also gives the possibility to provide citations for parts of the generated output.

\textbf{Cascaded Machine Translation (MT)} is included at the system level to enable support for multiple languages.
Specifically, non-English queries are first translated to English for our underlying LLM and retrieval engine to generate an English answer.
Finally, this answer is translated back to the original language for display.

\textbf{Benchmarking and Evaluation} is done both automatically and with human experts.
For automatic evaluation, we evaluate the model both on climate-domain tasks and general-domain tasks.
Furthermore, we describe our approach to human evaluation with domain experts, hoping to address the limitations that come with the automatic evaluation of LLMs.
We present the results of an initial human evaluation comparing our main model variants.

\textbf{Responsible AI} is an important aspect of our work because as LLMs become stronger, we strongly believe that the models should "do good."
To this end, we include instruction fine-tuning data that teaches the model to avoid answering unwanted or even malicious user queries.
During the whole development process, we carefully considered and actively worked on reducing the environmental impact of our work.
Finally, the models and evaluation protocols are released publicly to increase the reproducibility of our work.

\begin{table}
\begin{center}
\begin{tabular}{lrrrr} \toprule
Model & Base Model  & Tokens & LR & GPU Hours \\ \midrule
ClimateGPT-70B&Llama-2 70B & 4.2B & $1 \cdot 10^{-5}$ & 2,182 \\ 
ClimateGPT-13B&Llama-2 13B & 4.2B & $1 \cdot 10^{-5}$ & 301 \\
ClimateGPT-7B&Llama-2 7B & 4.2B & $1 \cdot 10^{-5}$ & 157 \\
ClimateGPT-FSC-7B  & -  & 319.5B & $3\cdot 10^{-4}$ & 14,131 \\
ClimateGPT-FSG-7B  & -  & 323.7B & $3\cdot 10^{-4}$ & 14,288 \\
\bottomrule
\end{tabular}
\end{center}
    \caption{ClimateGPT model variants.}
    \label{tab:model_variants}
\end{table}

\section{Domain-Specific Pre-Training}
\label{sec:pre_training}

Foundation models are pre-trained on vast datasets encompassing a wide array of domains \cite{brown2020language, touvron2023llama, touvron2023llama2}.
These domains range from general knowledge and common sense reasoning to more specialized areas like science, technology, and literature \cite{gao2020pile, penedo2023refinedweb}.
Training on a large-scale dataset enables the models to exhibit impressive zero-shot and few-shot (in-context learning) learning capabilities \cite{brown2020language, kojima2022large, wei2021finetuned, wei2022emergent}, allowing them to perform reasonably well on tasks they are not explicitly trained for.
However, despite their versatility, foundation models are not intrinsically designed to possess deep expertise in specific domains.
Therefore, recent efforts focused on training domain-specific language models that are either significantly smaller or outperform their general domain counterparts on domains like finance \cite{Wu2023bloombergGPT}, science \cite{taylor2022galactica} or medicine \cite{singhal2023expertlevel, chen2023meditron70b}.
To create such a model one can either perform domain adaptation on an existing general domain model \cite{singhal2023expertlevel, chen2023meditron70b} or train a new model from scratch \cite{Wu2023bloombergGPT, taylor2022galactica}.
Which approach is preferable depends on various factors, like the total compute budget, the amount of available domain-specific pre-training data and how well the target domain is represented in the general domain data.
To gain insights into these tradeoffs for the climate change domain, we compare both approaches.

In this section, we first describe the general model architecture we used for ClimateGPT.
Then, we describe how we curated and collected our climate change and science-specific pre-training dataset.
Next, we make use of continued pre-training as a domain adaptation technique to adapt a strong general domain model to the target domain.
Finally, we describe how we train a climate-specific model from scratch.

\subsection{Model Architecture}

We follow Llama-2 \cite{touvron2023llama2} closely in terms of the model architecture. Specifically, the model is a decoder-only Transformer \cite{vaswani2017attention,j.2018generating} network with word embedding layers sandwiching a stack of self-attention layers.
Key components, such as pre-normalization \cite{xiong2020layer} with RMSNorm \cite{zhang2019root}, SwiGLU \cite{shazeer2020glu} activation function, and rotary positional embeddings (RoPE) \cite{su2023roformer} are retained in this work.
Improvements on top of Llama-1 \cite{touvron2023llama}, such as increased context length (4096) and the introduction of grouped-query attention (GQA) \cite{ainslie2023gqa} for larger model variants were also kept.

\subsection{Pre-Training Dataset}
\label{sec:pretraining_data}

\begin{table}
\begin{center}
\begin{tabular}{lrr|rr|rr} \toprule
Subset & Tokens & Weight & \multicolumn{2}{c|}{Tokens in model} & \multicolumn{2}{c}{Percentage of data}\\ 
    &&&FSG&FSC&FSG&FSC\\
    \midrule
        news                 & 193.9 & 1 & 125.1 & 120.0 & 39.1\% & 37.5\% \\
        publications         & 23.1  & 4 & 59.6 & 57.1 & 18.6\% & 17.9\% \\
        modern\ books        & 28.4  & 3 & 55.0 & 52.7 & 17.2\% & 16.5\% \\
        patents              & 19.5  & 4 & 50.2 & 48.1 & 15.7\% & 15.1\% \\
        wikipedia            & 6.3   & 5 & 20.4 & 19.6 & 6.4\% & 6.1\% \\
        policy\ and\ finance & 3.7   & 3 & 7.1 & 6.8 & 2.2\% & 2.1\% \\
        science              & 0.7   & 5 & 2.2 & 2.1 & 0.7\% & 0.6\% \\
        \midrule
        climate change & 4.2 & 5 & 0 & 13.0 & 0.0\% & 4.1\% \\\midrule
\textbf{Total} & 279.7 & - & 319.5 & 100\% & 319.5 & 100\% \\ \bottomrule
\end{tabular}
\end{center}
    \caption{Subset breakdown of the 300B-token from scratch pre-training dataset.}
    \label{tab:from_scratch_data}
\end{table}

The preparation of high-quality in-domain data is important for the success of the model.
Therefore, we started with a corpus of roughly 300B tokens from curated sources compiled by Erasmus.AI.
While the corpus spans a wide range of domains sources were evaluated and selected based on their relevance to the topic climate, humanitarian issues and science.
The upper part of \Cref{tab:from_scratch_data} shows the different subsets of the dataset and the corresponding weight for model training.
The last columns indicate the effective number of tokens each of the from-scratch models has seen during from-scratch pre-training and the resulting data distribution (see \Cref{sec:from_scratch}).

The \emph{news} subset is a web crawl with a focus on relevant news and blog articles.
It also contains data from an internal extreme weather index.
\emph{Publications} is a collection of abstract and full-text papers.
The \emph{Modern books} set consists of fiction and non-fiction books and should help to model long-range context.
\emph{Patents} are collected mostly from the United States Patent and Trademark Office.
\emph{Wikipedia} is a recent dump of the English Wikipedia website.
\emph{Policy and finance} is a collection of text related to law, finance and companies and stocks in the climate sector.
Finally, \emph{science} covers other science and climate-related texts like EPA documents and ESG reports.

From this dataset, we identified high-quality sources such as scientific papers, and further included primary sources, such as reports from the Intergovernmental Panel on Climate Change (IPCC), and applied cleaning and filtering using keywords and topic classification.
In addition, we included our manually curated climate-specific data.
More details on these datasets are described in \Cref{sec:appendix_climate_data_details}.
In total, we arrive at a corpus of 4.2B climate-specific tokens which is used for continued pre-training.

To improve the quality of the training data, a set of cleaning, filtering, and pre-processing steps was done, which included:
\begin{itemize}
     \item filtering of sources from unrelated domains, such as sport and entertainment, politics and crime, as well as fiction. By doing so, we hope to limit the number of opinion pieces and information irrelevant to the climate domain;
    \item personal identifiable information reduction, such as email addresses, telephone numbers, URLs etc.;
    \item keeping sentences with a Flesch reading score \cite{kincaid1975derivation} between 5 and 120;
    \item handling of errors related to character encoding and special symbols;
    \item elimination of documents by text length;
    \item focusing on sources from the past eight years;
    \item aggregating themes concerning climate, humanitarian issues and science;
    \item discovering key sentences and entities that are associated with climate;
    \item filtering based on symbol distribution, i.e. removing documents that contain at least 80\% non-symbols;
    \item filtering based on language identification, i.e. removing documents that do not score above 85\% to be in English;
    \item removing double spaces, consecutive empty newlines, lines containing long repeating characters such as \enquote{======} and \enquote{+++++}, etc.;
    \item deduplication based on MinHash \cite{minhash} with proprietary extensions by Erasmus.AI as well as removal of duplication identifiable in the source metadata.
\end{itemize}

\subsection{Continued Pre-Training}
\label{sec:cpt}

\begin{figure}
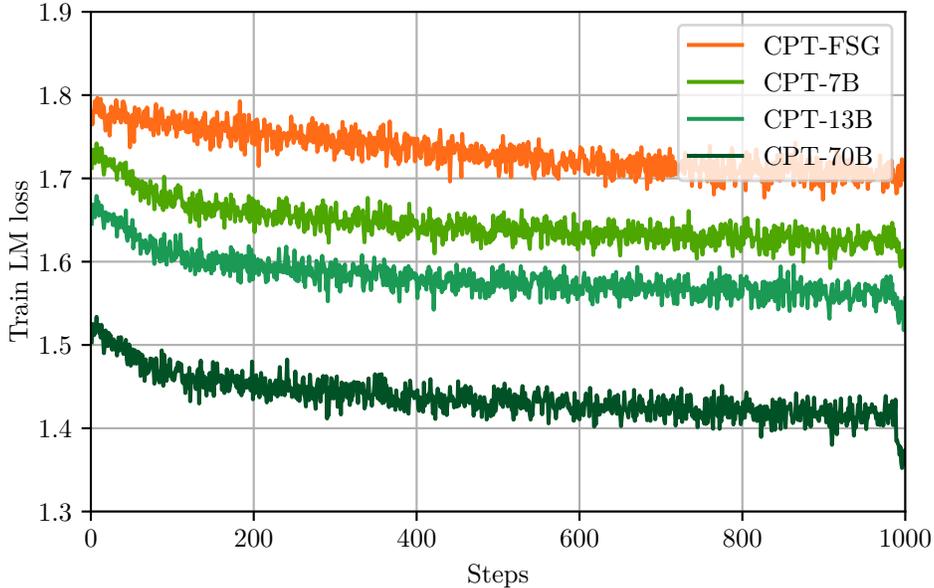

    \centering
    \include{figures/cpt.pgf}
    \vspace{-2em}
    \caption{Training loss of the CPT models.}
    \label{fig:cpt_training_curves}
\end{figure}

We employ domain adaptation methods, as we aim to develop a model that is specialized for climate change and possesses understanding and domain-specific knowledge. Domain adaptation can help tailor the foundation model to the climate domain, ensuring that it not only retains its broad knowledge base but also develops a more refined understanding of climate-specific concepts, terminologies, and contextual nuances.

Domain adaptation \cite{ben2010theory}, while not new, remains a cornerstone in the evolution of machine learning systems, e.g. in language modeling \cite{karouzos2021udalm}, machine translation \cite{kim-etal-2019-effective}, and automatic speech recognition \cite{baevski2020wav2vec}.
Fundamentally, the method involves the continued training of a baseline model on specific, in-domain data to enhance its performance within that domain.
This approach has been widely recognized for its ability to significantly boost a model's proficiency on in-domain test data, while still maintaining robust performance on general tasks.
In the context of our work, we adopt this principle to further refine foundation models for climate change applications.
We prefer to term this process as \enquote{continued pre-training} (CPT), rather than the more commonly used \enquote{fine-tuning}, to differentiate our approach from other methods like \enquote{supervised fine-tuning} \cite{ouyang2022training} and to highlight the similarity to the initial pre-training stage.
We deliberately apply this CPT step before proceeding to instruction fine-tuning.
If instruction fine-tuning is done before domain adaptation, there is a risk that the model might lose some of its newly acquired instruction-following capabilities.
By first adapting the model to the intricacies of the climate domain, we lay a solid foundation upon which instruction fine-tuning can then be built, ensuring a more effective and domain-savvy instruction-following model.

Domain adaptation, while offering significant benefits, presents two primary challenges.
The first challenge lies in preparing high-quality data for the in-domain training.
The effectiveness of domain adaptation is largely contingent on how closely the distribution of this training data aligns with that of the in-domain test data.
The closer the match, the more we can anticipate enhanced performance in domain-specific tasks.
To address this we make use of the curated dataset described in \Cref{sec:pretraining_data}.
The second challenge is the prevention of degradation in the model's performance on general domain tasks, a phenomenon often referred to as ``catastrophic forgetting" \cite{kirkpatrick2017overcoming} in the literature.
This occurs when a model, upon being further trained on specific data, loses its proficiency in tasks it was previously capable of handling.
To mitigate this, we carefully tune the batch size, learning rate, learning rate schedule, and data composition for the 7B model variants.
Due to time and compute-budget constraints during the project, we did not have time to tune the hyperparameters for the 13B and 70B models and just chose the same values as for the 7B models.

To choose a foundation model, we considered multiple candidates and chose the one that performed best on our climate-specific benchmarks (discussed in \Cref{sec:climate_specific_benchmarks}).
Candidates that we considered were Llama-2 \cite{touvron2023llama2}, Falcon \cite{almazrouei2023falcon}, Pythia \cite{biderman2023pythia} and Jais \cite{sengupta2023jais}.
From these models, we achieved the best results with Llama-2 (see Tables~\ref{tab:climate_automatic_results} and \ref{tab:general_automatic_results}), and thus we continued this model.
Redoing these experiments today, we would also consider Mistral-7B \cite{jiang2023mistral} and Mixtral \cite{jiang2024mixtral}, but these models were not available at this time.

For training, we use a fork of NVIDIA's Megatron-LM \cite{nvidia_megatron} by the EPFL LLM Team \cite{epfmgtrn, chen2023meditron70b}.
The main modifications to the original version from Nvidia are support for Llama and other recent models.
We use a cosine learning rate schedule with a peak learning rate of $10^{-5}$, a warm-up of 100 steps and decay to a learning rate of $5 \cdot 10^{-6}$.
The batch size is set to 1024 and we use the full sequence length of 4096 tokens.
For regularization, we use weight decay of $10^{-2}$.
All models are trained for 1,000 steps which corresponds to one epoch on the 4.2B climate dataset.
The training loss curves for the models are shown in \Cref{fig:cpt_training_curves}.

While we observed that higher learning rates resulted in better training and validation losses, we observed a degradation on our downstream benchmarks.
Therefore, we settled with this learning rate as a trade-off between domain adaptation and avoiding overfitting.

\subsection{From-Scratch Pre-Training}
\label{sec:from_scratch}

\begin{figure}
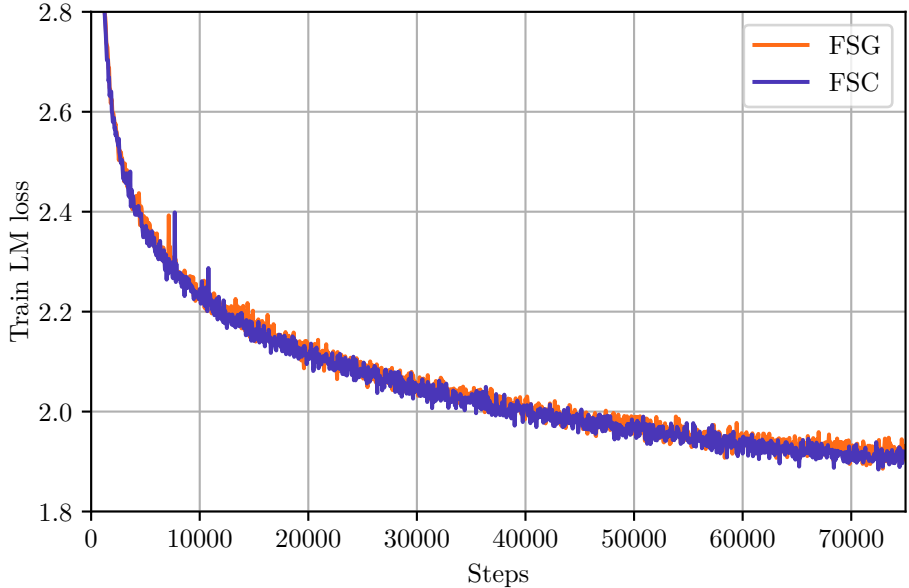

    \centering
    \include{figures/fs.pgf}
    \vspace{-2em}
    \caption{Training loss of the from-scratch general (orange) and from-scratch climate (purple) models.}
    \label{fig:fs_training_curves}
\end{figure}

In contrast to the continued pre-training approach, an alternative strategy involves departing from the use of pre-trained models such as Llama-2 \cite{touvron2023llama} or Falcon \cite{penedo2023refinedweb}.
Instead, we initiate the weights of a domain-specific foundation model entirely from scratch and directly train it on domain-specific data.

Adopting the approach of training a model from scratch comes with two significant implications. On the one hand, by choosing not to utilize a pre-trained foundation model, we inherently forego the advantages that come from training on the vast corpus of trillions of tokens that such models have been exposed to. These pre-trained models, despite not having fully disclosed datasets, are likely to have been trained on a diverse range of information, some of which could be beneficial for our purposes.
On the other hand, initializing the model from scratch offers us complete control over the training data, which is particularly crucial in a field like climate change that is prone to misinformation and bias \cite{coan2021computer}. By carefully selecting and curating the data, we can ensure that the model is trained on accurate, reliable, and scientifically valid information. This level of control allows us to mitigate the risk of perpetuating biases or inaccuracies that might be present in larger, less controlled datasets.
While we can expect better performance from training on more data \cite{kaplan2020scaling, hoffmann2022an}, projects developing domain-specific models often have lower compute budgets compared to general-purpose models with a broader range of applications.
Thus, training on less but more relevant and high-quality data could still result in better performance.

In our setup, we align our model architecture closely with the Llama-2-7B model developed by Meta, also utilizing the Llama tokenizer \cite{touvron2023llama}, which employs the Byte Pair Encoding (BPE) algorithm \cite{sennrich2015neural}.
We recognize that developing our own tokenizer, tailored specifically to climate-related terminology, could potentially give better vocabulary compression for domain-specific terms.
However, due to time constraints within the project, we left this for future work.
Nonetheless, our experience here provides a data point to judge the impact of different training datasets on model performance, while keeping other variables constant. For this reason, we continue with the rest of the development steps, such as instruction fine-tuning, with the from-scratch pre-trained model.

For from-scratch training, we use the same setup as for CPT training.
We use a cosine learning rate schedule with a peak learning rate of $3 \cdot 10^{-4}$, a warm-up of 100 steps and decay to $10\%$ of the peak learning rate, i.e. to $3 \cdot 10^{-5}$.
The batch size is set to 1040 and we use the full sequence length of 4096 tokens.
For regularization, we use weight decay of $10^{-1}$.
Both models are trained for 75,000 steps the resulting effective tokens seen per subset are shown in \Cref{tab:from_scratch_data}.
The training loss curves for the models are shown in \Cref{fig:fs_training_curves}.
To train both models we use the Adam optimizer \cite{KingBa15adam} with $\beta_1 = 0.9$, $\beta_2 = 0.95$ and $\epsilon = 10^{-5}$.
While these values are commonly used to train large language models \cite{brown2020language, biderman2023pythia, touvron2023llama2}, we want to highlight that decreasing the $\beta_2$ momentum from the common default value of $0.999$ decreases training instabilities and loss spikes caused by large batch sizes \cite{zhai2023sigmoid}.

\subsection{Training Hardware}

Given the computationally intensive nature of training a foundational model from scratch \cite{hoffmann2022an}, there are significant environmental considerations, especially pertinent in the context of our work in the climate domain. Therefore, we choose to utilize a high-performance computing cluster that is entirely powered by hydropower (24g CO$_2$eq / kWH \cite{ipcc2014hydro}), provided by MLFoundry. The cluster has 32 nodes, each equipped with 8 H100-SXM GPUs. These nodes are interconnected through InfiniBand, ensuring high-speed data transfer and communication across nodes. Additionally, within each node, the GPUs are connected via NvLink, facilitating efficient intra-node GPU communications. Leveraging Megatron's \cite{shoeybi2019megatron, epfmgtrn} efficient implementations of data parallelism, tensor parallelism, and pipeline parallelism, we achieved an average training speed of 250 TFLOPS per GPU, and the training run took 3.7 days on 20 nodes.
When fully utilized, we assume a power consumption of 775W per GPU (including CPU).

\section{Instruction Fine-Tuning}
\label{sec:ift}

\begin{figure}
    \centering
    \includegraphics[width=\linewidth]{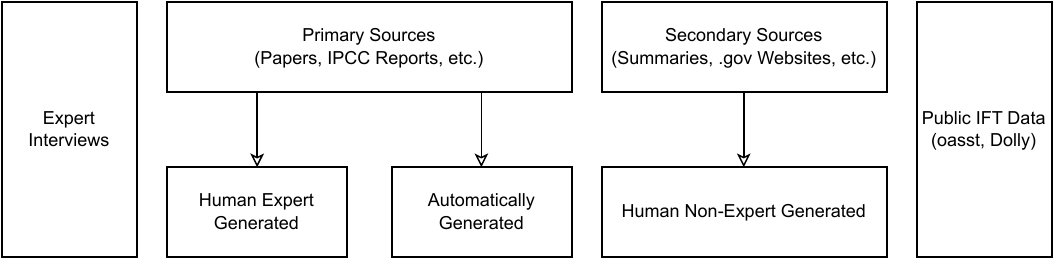}
    \caption{Instruction Fine-Tuning Tracks}
    \label{fig:ift_tracks}
\end{figure}

After pre-training, we expect that the resulting domain-specific language models have a deeper understanding and knowledge of the target domain than comparable foundation models.
Since these models were merely trained to predict the next token in our pre-training dataset, using them for specific downstream tasks requires careful prompting or providing the model with few-shot examples.
Adapting these models to follow users' instructions formulated in natural language and generate answers in a style appropriate for our use case requires Instruction Fine-Tuning (IFT) \cite{ouyang2022training}.
In the literature, this is also often referred to as Supervised Fine-Tuning (SFT), but we use this terminology as a clearer distinction to other fine-tuning steps (e.g. CPT or task-specific fine-tuning).
To do this, the model is trained on instruction and corresponding completion pairs.
In the following, we are also referring to these pairs as demonstrations.
To limit the complexity of the required data, we decided to only focus on prompt and completion pairs and not to collect any multi-turn chat interactions.

To adapt the style of completions to be appropriate for our envisioned use case, we require a sufficiently large amount of in-domain data for IFT.
However, collecting such a dataset is challenging, as it requires a certain level of expertise in the target domain.
During the project, we had the privilege to be able to work with a small team of climate experts as well as a larger team of non-experts with limited domain knowledge.

\Cref{fig:ift_tracks} shows the different tracks we followed to collect IFT data.
The first track of our IFT data consists of demonstrations (i.e. instruction and completion pairs) collected through \emph{interviews with senior climate experts} (i.e. experienced researchers in the field like professors or other leading experts).
During the interviews, the main questions in the field of study of the expert were discussed, implications on other fields as well as different use cases for a climate-specific LLM.

For the remainder of the collected climate-specific IFT data, we provided annotators with existing documents as the basis for the demonstrations they generate.
We identified that coming up with new topics can be a limiting factor in the data creation process for the annotators, and this approach can help to concentrate their mental load to write a good completion.
Additionally, having control over these \enquote{seed} documents means that we can increase the diversity of topics to be covered in the IFT corpus.
For the \emph{human expert generated} part of our IFT data, non-senior climate experts (i.e. graduate or PhD students or other early career researchers) created data based on primary sources (like research papers and technical sections of the IPCC reports).
As the time of climate experts is valuable and limited we additionally worked with a larger team of non-expert data annotators.
Most primary sources, like climate change papers or technical sections of the IPCC reports, were not completely comprehensible by the non-experts.
Therefore, for this team, we decided to focus on secondary sources such as governmental websites (e.g. from the EPA, NASA, or European Parliament) and summary sections of the IPCC reports.
As we were still concerned that we might not be able to collect a sufficient amount of in-domain data, we simultaneously experimented with \emph{synthetically generated} demonstrations from the documents using a general-purpose LLM as opposed to the manual IFT creation from above.

Finally, to increase the amount and diversity of instruction-tuning data and to be able to reuse well-developed non-climate domain-specific instructions, we made use of existing \emph{general domain IFT data}.
\Cref{tab:ift_data} gives an overview of the different IFT datasets that were used to train ClimateGPT.
The rest of this section is devoted to providing more details on these datasets and how they are used to train ClimateGPT.

\begin{table}
\begin{center}
\begin{tabular}{llrrr} \toprule
 Domain & Name & Total Size & Training Samples & \\ \midrule
Climate& Senior Expert Interviews & 74 & 1,332 & \\
 &Grounded Expert Demonstration & 403& 7,254 & \\
 &Grounded Non-Expert Demonstrations & 9,663 & 146,871 & \\
 &Synthetically Generated Demonstrations & 57,609& 0 \\ 
 &StackExchange & 3,282 & 9,846 & \\\midrule
General &AppTek General &700 & 2,100 & \\
& OASST-1 & 3,783 & 11,349 & \\
&Dolly & 15,001 & 45,003 & \\
&Llama-2 Safety & 939 & 2,817 & \\
 &FLAN & 38,909 & 30,000 & \\
 &CoT & 448,439 & 15,000 & \\
 \midrule
 Total& & & 271,572 &  \\ \bottomrule
\end{tabular}
\end{center}
    \caption{Details about the Instruction Fine-Tuning datasets.}
    \label{tab:ift_data}
\end{table}

\subsection{Senior Expert Interview Demonstrations}
\label{sec:ift_interviews}

Our vision for the model is for everyone to have a personal climate expert at their fingertips breaking down questions and concepts to the level of expertise of the user.
Interviews with climate experts most closely resemble this goal and thus IFT data created in this process is the most valuable data source for ClimateGPT.
We started the interviews by defining foundational concepts in the area of expertise of the interviewee and the role of climate change.
Second, we discussed current trends in the field and the expected developments in the future.
Next, we discussed pivotal findings and research papers in the field and extracted key arguments.
Finally, we brainstorm ways in which a climate-specific LLM could be helpful for stakeholders involved in this specific field.
As the time of the corresponding experts is very limited, instruction and completion pairs were developed afterward by the interviewer.

For the first version of ClimateGPT, we conducted a series of interviews with the agricultural ecologist Dr. David Lobell.
He is the Director of the Center on Food Security and the Environment at Stanford University and also served as lead author for the food chapter on the IPCC Fifth Assessment Report (AR5).
The result of this process was a high-quality IFT dataset of 74 demonstrations.
Based on these promising results, we plan to refine our methodology and conduct additional interviews.

\subsection{Grounded Expert Demonstrations}
\label{sec:ift_experts}

In addition to the non-expert annotators, we collaborated with nine climate scientists (graduate or PhD level) from different European universities.
For the data collection, AppTek's data annotation tool Workbench\footnote{\url{https://www.apptek.com/technology/workbench-data-annotation}} was used.
The team worked in close collaboration with the authors to improve the style of the generated data.

As a first step, we asked the nine climate scientists to think themselves about five to ten questions relevant to climate change that they find important to address and feel comfortable answering.
We proposed to them to organize their answers with a scientific mindset (the style we want ClimateGPT to use) by first making a summarizing statement followed by a list of bullet points explaining or developing elements of the summary.
Each answer should refer to a scientific source, from which the experts should extract a couple of paragraphs relevant to the answer.
The retrieved paragraphs were stored so that they could be used later on.
This first exercise was a first test used to evaluate the writing skills of our experts. At the end of this first phase, we continued with seven of the nine experts with the IFT creation task.

In the second step, we provided the experts with references to primary sources and questions related to these sources that have been generated by our synthetic IFT pipeline (\Cref{sec:apex}), as a source of inspiration.
The synthetically generated questions can be very specific reading comprehension questions with respect to the reference source that often has no relevance outside of the source document used. Such questions would either need to be generalized or skipped by the expert. Also, to be time-efficient while producing an answer, we proposed to our experts to choose those questions that relate to their domain of expertise (e.g. city climate, tropical climate, etc.). At the end, we gave each expert a set of 1,000 question-answer pairs, from which 50 to 250 have been selected.
In contrast to the non-expert data collection effort (\Cref{sec:non_expert_data}), we did not suggest specific task categories to the experts and instead let them decide on relevant instructions.

As addressed previously, we want the model to be able to generate different in-depth responses addressing the different dimensions of climate change, namely natural science, economics, and social aspects.
To collect IFT data for this feature, we asked the expert annotators to create four responses to the same prompt one giving a general answer and three focusing on one of these dimensions.

\subsection{Grounded Non-Expert Demonstrations}
\label{sec:non_expert_data}

\begin{table}
\begin{center}
\begin{tabular}{lrr} \toprule
Task Category & \% \\ \midrule
Open Ended QA & 26.9 \\
Open Ended Generate & 48.0 \\
Open Ended Classification & 1.2 \\
Open Ended Chat & 4.7 \\
Open Ended Chain of Thought & 0.1 \\
Open Ended Brainstorm & 7.5 \\
Closed Ended Summarize & 4.7 \\
Closed Ended Rewrite & 0.2 \\
Closed Ended QA & 3.8 \\
Closed Ended Extract & 1.6 \\
Closed Ended Classification & 1.3 \\ \bottomrule
\end{tabular}
\end{center}
    \caption{Task distribution for the non-expert data collection.}
    \label{tab:task_distribution}
\end{table}

For the non-expert data collection, we worked with a team of 99 annotators employed by external contractors from six different countries and three continents.
Annotators were selected based on their educational background, domain-specific expertise and interests, strong communication skills, and writing skills.
More details on the demographics of the annotators are provided in \Cref{sec:annotator_demographic}.
All annotators were trained by the corresponding project managers on the project scope, guidelines and requirements.
The team used the same tool as the expert annotators (\Cref{sec:ift_experts}).

To ensure a certain level of diversity of types of instruction, we provided annotators with a task category.
The set of tasks and their distribution is based on the use case categories reported in \cite{DBLP:conf/nips/Ouyang0JAWMZASR22}.
The resulting task distribution is shown in \Cref{tab:task_distribution}.
For each category textual guidelines were provided to the annotators.

For the initial phase of the project, we did not provide annotators with any specific topic to work on in addition to the general climate topic.
However, we observed that this resulted in too many simplistic and overlapping prompt and completion pairs.
Providing annotators with a specific topic to work on resulted in more diverse and interesting data.
Topics were selected based on interviews with climate experts to cover the climate impact across various real-life situations and elements.
\Cref{tab:topic_distribution} in \Cref{sec:annotator_demographic} shows the full list and distribution of topics.

The recommended way of data creation was to find content from approved data sources to develop ideas for prompt and completion pairs.
Initially, we provided annotators with primary sources, such as research papers and technical sections of the IPCC reports.
However, initial feedback showed that our annotators struggled with these documents.
Therefore, we decided to switch to secondary sources, such as governmental websites (e.g. from the EPA, NASA, or the European Parliament) and summary sections of the IPCC reports.
Besides the trustworthiness of the content, data sources were approved to avoid copyright issues.

Annotators were instructed to give in-text citations to sources they were using in the completion.
We instructed annotators to give citations in MLA style (i.e. author name, title, and source in brackets) but noticed that this resulted in inconsistencies that had to be corrected in post-processing.
Later we switched to IEEE style (i.e. reference number in square brackets).
The data annotation tool allows storing additional details for each citation, such as the URL or the cited text, as additional metadata for the prompt-completion pair.
At the beginning of the data annotation process, we decided to instruct annotators to only store the URL of the cited source and not the cited text itself.
While the latter would have been useful to improve the retrieval augmented generation capabilities of the model, we decided against it in concern that annotators would restrict themselves to the referenced text (instead of making use of all information in the document) and to avoid increasing the complexity of the annotations process, and, thus, the volume of data we can collect.
As an alternative, we can make use of the URLs to crawl the complete document and reconstruct the cited paragraph automatically.
\Cref{sec:attribution} discusses this process in more detail.

\subsection{Synthetically Generated Demonstrations}
\label{sec:apex}

As access to experts who can make use of primary sources is limited (and we initially were concerned that we may not be able to collect enough human-generated IFT data), we were also investigating synthetically generating demonstrations from primary sources.
To achieve that we prompted an existing general-purpose LLM with few-shot examples and a document and instructed the model to first generate a question and the corresponding system completion.
The prompts were carefully designed to increase the diversity of the generated data.
Further, we applied multiple post-processing steps to ensure that the generated data is of high quality.
These steps included verifying that there is not too much and not too little overlap to the reference documents and prompting general-purpose LLM again to check whether the generated completion is plausible.
Further, we filter out questions or responses that mention figures or specific sections from papers and try to detect other text generation artifacts like repeating sequences.
This process was initially designed with a multi-turn model in mind.
Therefore, completions were intentionally kept shorter with the intent that the user might ask follow-up questions.
The decision not to allow multi-turn interactions in this initial version and that more comprehensive answers are preferable came later in the project.

While initial experiments showed promising results, we did not observe consistent improvements in our automatic benchmarks for later versions of the models when using this data.
Thus, due to this and due to the lenght mismatch this data is not directly included in our final IFT data mixture.

\subsection{General Domain Data}
\label{sec:ift_general}

As the last track of our IFT training dataset, we make use of existing human-written IFT datasets that are available to us.
The first is an internal high-quality set of prompt-completion pairs originally collected by AppTek.
We are referring to this dataset as \emph{AppTek General}.
The data collection methodology was similar to the one described for the non-expert data collection.

Further, we make use of two openly available crowd-sourced IFT datasets.
First, \emph{Databricks Dolly} \cite{DatabricksBlog2023DollyV2} was the first openly available human-generated IFT dataset with a permissive license.
The dataset consists of 15,001 prompt and completion pairs across 7 task categories and was generated over two months by over 5,000 employees at Databricks.
Second, \emph{OpenAssistant Conversations 1} (OASST-1) \cite{koepf2023openassistant} is a dataset consisting of 161,443 messages in 35 different languages.
The corpus is the result of a worldwide multilingual crowd-sourcing effort involving over 13,500 annotators.
In contrast to all previously mentioned IFT datasets, this dataset does not only contain instruction and completion pairs but also multi-turn conversations.
For ClimateGPT, we only make use of English conversation and only include the best-rated messages in each conversation tree, resulting in a total of 3,783 conversations.

As an additional source of data, we included 3,282 question-and-answer pairs from domain-relevant \emph{StackExchange} communities (earth science, sustainability and economics).
Another common approach to curating IFT datasets is to format existing NLP datasets as instruction and completion pairs using task-specific templates \cite{wang-etal-2022-super, longpre2023flan}.
While training on this type of data alone is not sufficient to achieve good performance \cite{DBLP:conf/nips/Ouyang0JAWMZASR22}, combining this type of data can be beneficial \cite{wang2023how}.
A possible explanation for this is that this way the model is exposed to a larger variety of tasks and more examples of in-context learning.
At the same time, this type of data is closer to our evaluation tasks than human-written pairs, which might explain improvements in automatic evaluation that might not translate to improvements under realistic use.
We decided to include 15,000 examples per epoch from \emph{FLAN v2 and CoT} as described by \citet{wang2023how} into our training data.

Most recently published instruction fine-tuning datasets were created by distillation from large proprietary LLMs like GPT-4.
Examples of these include Alpaca \cite{alpaca}, Vicuna \cite{vicuna2023} or WizardLM \cite{xu2023wizardlm}.
We intentionally decided not to make use of these datasets.
First, recent research has shown that training on these approaches can successfully transfer the style of the models but not their factuality \cite{gudibande2023false}.
Second, the biases of the teacher model may be transferred to the student.
And finally, the licensing terms of commercial LLM providers often limit the use of their API to train models that potentially compete with them.
Due to this, the usage of this type of data in models intended for commercial use is problematic \cite{alpaca}.

\subsection{Safety Data}

One missing component in our IFT dataset is examples to align the model to be safe and harmless.
While both Dolly and OASST-1 contain a few examples of refusing to answer intentionally harmful prompts, we observed that this was not enough to make the model safe.
To evaluate this, we analyzed completions of initial versions of the model on a subset of the Do-Not-Answer dataset \cite{wang2023donotanswer}.
This dataset consists of around 1,000 prompts that are intentionally designed to invoke harmful or offensive model outputs.
As expected, the initial model produced multiple unsafe and potentially harmful outputs, which suggests that additional demonstrations of expected model behavior are required.
As writing safe completions to these types of prompts can be especially stressful for annotators and new approaches to safety are not the center of this work, we decided to make use of an already safe model to generate the completions synthetically.
Specifically, we generated completions for each prompt in the dataset using Llama-2-Chat-70B \cite{touvron2023llama} and included this data in our IFT set.
We are referring to this dataset as \emph{Llama-2 Safety}.
The design considerations around safety are discussed in more detail in \Cref{sec:content_moderation}.

\subsection{Data Preparation}

We use a mix of different sources for our IFT data to enable alignment with the different aspects outlined in the previous sections.
\Cref{tab:ift_data} shows the mixing ratios of the different subsets in our final model training.
We just train for a single epoch on the general domain data and up-sample the climate-specific data.

During inference, we want the model to generate text that is as close as possible to our expert-generated data.
However, since the majority of the IFT data comes from other sources, we need some mechanism to counteract this.
Our solution to this is to use different system prompts for each data source to condition the model.
By using the system prompt corresponding to the expert IFT data we can control its style at inference time.
Further, this also allows us to train on data that, e.g. does not make use of all the features of the model (e.g. does not provide citations to references as discussed in \Cref{sec:attribution}).
The system prompts for each of the subsets are listed in \Cref{sec:appendix_system_prompts}.

To prepare the IFT data for training, we make use of the codebase from Open Assistant\footnote{\url{https://github.com/laion-ai/open-assistant/tree/main/model/pretokenizer}}.
During training the IFT data is formatted using the ChatML prompt template\footnote{\url{https://github.com/openai/openai-python/blob/120d225b91a8453e15240a49fb1c6794d8119326/chatml.md}}, following other recent open source models like Open Assistant\footnote{\url{https://huggingface.co/OpenAssistant/llama2-70b-oasst-sft-v10}} or Meditron \cite{chen2023meditron70b}.
Standardizing prompt templates in open-source models results in greater compatibility with existing tools and libraries.

\subsection{Training}

\begin{figure}
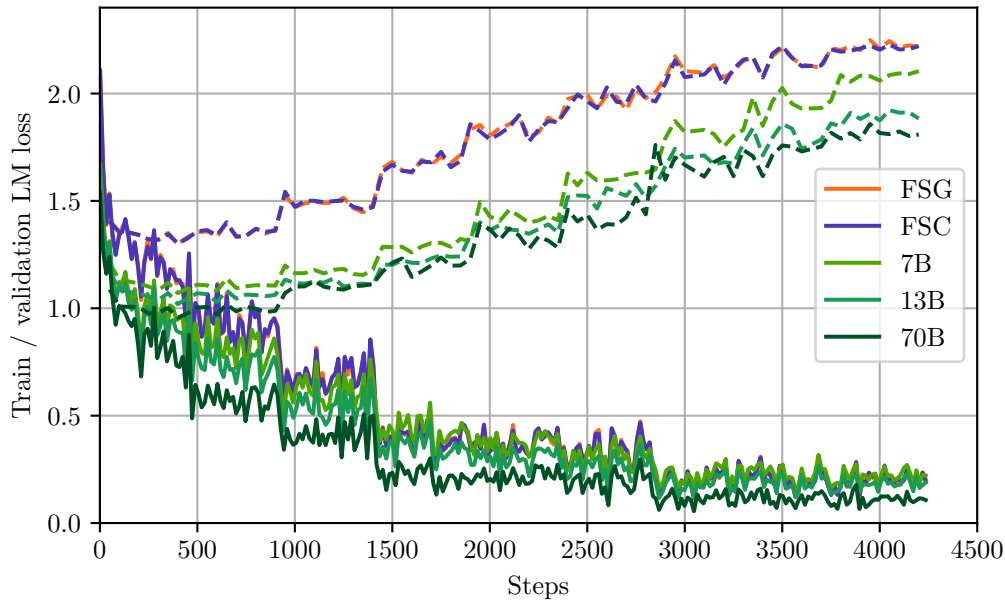

    \centering
    \include{figures/ift.pgf}
    \vspace{-2em}
    \caption{Training (solid) and validation (dashed) loss of the final IFT models.}
    \label{fig:ift_training_curves}
\end{figure}

As for pre-training, we use a fork of NVIDIA's Megatron-LM \cite{nvidia_megatron} by the EPFL LLM Team \cite{epfmgtrn, chen2023meditron70b} for IFT training.
We use a cosine learning rate schedule with a peak learning rate of $10^{-5}$ and a warm-up of 100 steps.
The batch size is set to 64 and we use the full sequence length of 4096 tokens.
For regularization, we use weight decay of $10^{-2}$ and dropout as used for LIMA \cite{zhou2023lima}.
The training and validation loss curves for our models are shown in \Cref{fig:ift_training_curves}.
As the validation set, we used a held-out set of 400 prompt and completion pairs from our non-expert climate data.

As was observed by previous work \cite{DBLP:conf/nips/Ouyang0JAWMZASR22, zhou2023lima}, the perplexity on the validation set first decreases for the first steps and then increases.
Typically, this is a clear sign of over-fitting, but, as in other works, we observe that the quality of the model still improves.
This is measured by evaluating the model on our automatic evaluation benchmark (\Cref{sec:automatic_evaluation}).

\section{Retrieval Augmented Generation}
\label{sec:rag}

While pre-training and instruction fine-tuning on climate-specific data improve the climate knowledge of the model, large language models still tend to hallucinate information, especially facts that are not well represented in the training data.
An example, where this is especially problematic, is specific numeric figures, for example, CO$_2$ emissions of a country in a specific year.
In addition, the model's knowledge is frozen and it is not possible to incorporate new facts or knowledge without additional training or other approaches to modify the model's weights \cite{pmlr-v162-mitchell22a}.
In the domain of climate change, these issues are critical.
The field is constantly evolving and having access to the latest findings is crucial to draw relevant conclusions.

Retrieval augmented generation (RAG) addresses both of these issues by retrieving relevant documents from external databases and providing these documents as additional context to the model.
While the approach was originally proposed for question-answering \cite{chen-etal-2017-reading,realm}, it has been successfully applied to other tasks like machine translation \cite{khandelwal2021nearest}, task-oriented dialog \cite{thulkeEfficientRetrievalAugmented2021} and recently in the context of instruction-tuned LLMs \cite{shi2023replug}.
The general approach for RAG is to have a separate retriever module which given the user query returns a list of relevant documents.
Then, in the language model -- often referred to as reader in this context -- both user query and retrieved documents are given as context to the model to produce a response.
Due to the limited sequence length of transformers and efficiency concerns, it is often not feasible to provide full documents (like full research papers) to the model.
Instead, shorter excerpts of a few sentences or paragraphs are typically used as the unit for retrieval and input to the model.
Systematic studies show that RAG can reduce the number of hallucinations in language models \cite{shuster-etal-2021-retrieval-augmentation}.

Nevertheless, the approach still suffers from limitations.
Models usually suffer from noise or irrelevant documents in the retrieval context \cite{pmlr-v202-shi23a, cho2023improving}.
We address this issue by including distractor documents during IFT training to allow the model to learn to ignore irrelevant documents.

While RAG is an obvious approach to increase the factuality of generated responses, it also suffers from an inherent trade-off between factuality and abstractiveness \cite{daheim-etal-2022-controllable, daheim2023elastic}.
With current RAG approaches, generated responses are often limited to the information provided in the retrieved documents and do not provide a broader perspective.
For climate communication, it's especially important to provide an interdisciplinary perspective integrating different viewpoints in the response.
To address this in our approach, we propose to make use of distinct sets of documents to generate different answers each covering one of the main perspectives.

We would like to note that RAG is widely used in literature to improve LLMs in the domain of climate change and communication
First, ChatClimate \cite{chatclimate} makes use of GPT-4 as LLM and follows the standard approach outlined above.
As a document source, the IPCC AR6 reports are used. The reports are converted to raw text and split into smaller chunks.
For retrieval, OpenAI's \texttt{text-embedding-ada-002}\footnote{\url{https://openai.com/blog/new-and-improved-embedding-model}} embedding model is used.
In contrast to our approach the retrieval database is just limited to IPCC reports and the base LLM was not adapted to the domain.
The system explicitly refuses to generate completions for prompts for which no relevant passages can be found in the IPCC reports.
Secondly, for Arabic Mini-ClimateGPT \cite{Mullappilly2023arabic} the authors fine-tune Vicuna-7B \cite{vicuna2023} which is based on the first version of Llama-7B \cite{touvron2023llama}.
Similar to this work, they dynamically retrieve both English and Arabic climate-specific documents but do not specify the source of the documents in more detail.
In contrast to us, they do not perform continued pre-training to adapt the model to their domain.
Furthermore, their IFT data was synthetically, generated using ChatGPT, while our climate-specific IFT data was manually curated by humans in close cooperation with multiple climate scientists and experts.

\subsection{Retrieval Dataset}
\label{sec:rag_dataset}

\begin{table}
    \begin{center}
    \begin{tabular}{lrr}
        \toprule
        Source & \# Docs & \# 512 Chunks \\ \midrule
        IPCC Reports & 16 & 17897 \\
        Potsdam Papers & 390 & 8539 \\
        Earth4All & 14 &  235 \\ 
        73 other (open access) & 336 & 8648 \\
        \bottomrule
    \end{tabular}
    \end{center}
    \caption{Statistics of the different data sources of the primary retrieval dataset.}
    \label{tab:retrieval_dataset}
\end{table}

The dataset for retrieval consists of a manually curated collection of scientific reports and papers.
We used the IPCC reports as a starting point and then manually extended the dataset with additional trusted sources in collaboration with climate experts.
Therefore, we focused on recent documents to avoid including outdated research.
During the data collection, we carefully evaluated the license of each document and only included content with open access or Creative Commons licenses allowing commercial use.
To reduce the complexity of the text extraction pipeline, we only considered digitally native PDF documents (i.e. documents where the content can be directly extracted without requiring OCR or similar approaches).

After collecting the PDF documents, we first split the documents into separate PDF pages and for each page, the text is extracted using PyMuPDF\footnote{\url{https://pymupdf.readthedocs.io/en/latest/}}.
While with this approach we might split relevant paragraphs in the middle of a sentence and loose cross-page context, it greatly simplifies our data processing pipeline.
Then, the text on each page is split into chunks of 115 tokens.
Next, we observed that many pages in these documents do not contain any relevant information for RAG and potentially degrade performance.
These include, for example, tables of content or pages with references.
These pages have a high density of superficially relevant content and thus are likely to be retrieved.
On the other hand, in most cases, these pages do not provide the full information required to generate a response.
To remove these pages, we use a combination of manual data cleaning as well as heuristics to detect problematic content. 
We deployed a custom tool to iterate, filter, and manually edit data and end with a final set of curated and clean data.
The resulting pages are converted to sub-word tokens and then split into chunks of length 115 with stride 10.
As the last step, we filter chunks that do not contain enough information (e.g., chunks only containing numbers from tables).

\subsection{Retrieval Approach}

For retrieval, we follow the common approach of using a transformer bi-encoder model \cite{mazare-etal-2018-training,reimers-gurevych-2019-sentence}.
Here, both document and query are passed separately through a transformer encoder to produce embedding vectors for both sequences of tokens.
The similarity between the query and the document is then measured by calculating the dot product or cosine similarity between corresponding vectors.
The main advantage of this approach is that embeddings for all documents can be pre-computed and only the query has to be passed through the model at inference time.
Other retrieval methods, like cross-encoders \cite{reimers-gurevych-2019-sentence}, pass query and document through the model simultaneously.
While this results in better retrieval performance, the inference cost becomes prohibitively expensive if the document database exceeds more than a few hundred documents; as in the case of our approach, where we want to access a broad range of content.
Therefore, the cross-encoder approach is commonly used only to re-rank results from other more efficient methods.

\begin{table}
    \begin{center}
    \begin{tabular}{lrrrrr} \toprule
        & & \multicolumn{2}{c}{Question} & \multicolumn{2}{c}{Answer} \\
    Model & Params & R@1 & R@5 & R@1 & R@5 \\ \midrule
    bge-base-en-v1.5 &0.1B& 54.8&	71.5  & 81.8&	92.1 \\
    bge-large-en-v1.5 &0.3B&  55.8 & 73.6 & 83.3 & 93.1 \\
    gtr-t5-large &0.3B&48.8	&67.4&79.6	&90.1 \\
    gtr-t5-xxl &4.8B&47.6	&66.3 &79.2	&89.7 \\
    gte-large&0.3B& 50.7	& 68.2 & 80.9	&91.4 \\
    ember-v1 &0.3B&49.5	& 68.6 & 79.7	&91.1 \\
    instructor-large &0.3B&50.0	& 68.2 & 81.7	&91.8 \\
    instructor-xl &1.2B&53.3	& 69.7 & 83.3	&92.1 \\ \bottomrule
    \end{tabular}
    \end{center}
        \caption{Recall (R@1 and R@5) of retrieving the correct document given the question or the answer from the synthetically generated IFT dataset.}
        \label{tab:retrieval_results}
\end{table}

As training our own retrieval model was out-of-scope for this project, we evaluated several existing embedding models.
We only considered bi-encoder models and did not integrate an additional model for reranking.
As an initial set of models, we considered the best-performing models on the MTEB benchmark \cite{muennighoff-etal-2023-mteb}.
To be able to do our own in-domain and use-case-specific evaluation, we selected a subset of the synthetically generated IFT data (\Cref{sec:apex}) as a test set.
The generated question and source paragraph are considered positive pairs and all other paragraphs as negatives.
We run the benchmark both with the question and with the answer as a query.
The results of this evaluation and the models under consideration are shown in \Cref{tab:retrieval_results}.
We got the best results using \texttt{bge-large-en-v1.5} \cite{bge_embedding} and decided to use this embedding model for retrieval.

To increase the relevance of retrieved chunks, we use a hierarchical retrieval approach.
Therefore, we first retrieve the top-$k$ PDF pages.
We use the full text on the page to calculate the embedding.
If the length of the text on a page exceeds the maximum sequence length of the embedding model (512 in our case with variable stride length), we use a sliding window over the sequence and average the embeddings to get a single embedding for the whole page.
After selecting the top-$k$ ($k=60$) pages, as a last step, we retrieve the top-$k$ ($k=5$) most relevant chunks corresponding to these pages (note that the number of chunks per page varies).
This ensures that the page context of retrieved chunks is also relevant to the query.

HuggingFace Transformers \cite{wolf-etal-2020-transformers} and Sentence Transformers\footnote{\url{https://github.com/UKPLab/sentence-transformers}} \cite{reimers-gurevych-2019-sentence} was used to deploy the embedding models and to embed text in preprocessing and inference stages.
For efficient nearest-neighbor search we use ScaNN\footnote{\url{https://github.com/google-research/google-research/tree/master/scann}} \cite{scann_paper}.

\subsection{Grounding}
\label{sec:attribution}

To improve the model's capabilities to make use of reference paragraphs provided in the context, we train the model with IFT examples that already include reference paragraphs.
For the IFT data collected from climate experts (\Cref{sec:ift_experts}) and from interviews (\Cref{sec:ift_interviews}) we asked the annotators to provide the reference paragraphs as additional metadata.
For the non-expert data (\Cref{sec:non_expert_data}), we used the URLs of the cited sources to crawl corresponding documents and extract cited paragraphs.
Most of the cited sources were websites; therefore, we constructed a separate pipeline to crawl these websites, extract the text using Mozilla's readability\footnote{\url{https://github.com/mozilla/readability}} and inscriptis\footnote{\url{https://github.com/weblyzard/inscriptis}}, and split the content into smaller chunks.

As also shown in \Cref{tab:retrieval_results}, it is easier to retrieve the correct paragraph using the completion as a query than using the prompt.
The reason for this is that the completion contains all the relevant information from the document and thus has high semantic similarity.
In contrast, the question just asks for the corresponding information which may not necessarily imply semantic similarity.
In our case, we can make use of this fact by not only considering the relevance of the prompt to the potential reference chunks, but also the relevance of the answer.
Furthermore, as discussed at the beginning of this section, numbers are a common source of hallucinations in LLMs.
Thus, we want to make sure, that the numbers that annotators extracted from the provided reference are part of the selected reference chunk.
To achieve this we add the overlap in numbers between the completion and potential reference paragraph as an additional scoring factor.
Based on these three scores, we select the best matching chunk from the reference document as source chunk which is given as context during IFT training.

To increase the robustness of the model to irrelevant retrieval results, we added distractor chunks as additional context.
These chunks are selected from the total set of chunks produced by the pipeline above.
To select chunks not relevant to the answer, we use the opposite of the scoring function outlined above, i.e. we select chunks with high similarity to the prompt but with low similarity to the answer and with low numerical overlap.
Additionally, some tasks for the model do not require information from any reference paragraph.
Examples include all closed-ended task categories described in \Cref{tab:task_distribution}.
For these tasks, the prompt already includes all relevant context, so everything that is retrieved is just unnecessary noise.
Instead of detecting this at inference time and disabling the retrieval augmentation, we also include distractor paragraphs during the training of IFT examples of these categories to make the model more robust.

\subsubsection{Citations}

In a scientific context, it is important that provided information is attributable.
In RAG, the retrieved chunks are attributable since their source is known.
However, the generated completion is not necessarily explicitly related to any of the chunks. 
It would be helpful to know which chunks exactly were used to generate which part of the response.
As discussed in \Cref{sec:non_expert_data}, annotators were asked to provide in-text citations for each reference.
We converted these citations to special tokens \texttt{[[0]]}, \texttt{[[1]]}, etc. which are prepended to the corresponding chunk in the context as well as the token for the citation.
While the general approach seems to work, we observed that citation tokens are hallucinated in some cases.
We attribute this to the automatic extraction of reference chunks that might not cover all relevant details of the source documents and, thus, introduce noise.
For this reason, we removed citations from the IFT data for the final ClimateGPT models and plan to reconsider this in the future.

\subsection{Three Dimensions}

Climate change is inherently an interdisciplinary field.
Therefore, to effectively serve our intended audience, including policymakers, scientists, and journalists, we aim to enhance our model to adeptly address queries from three critical perspectives: natural, economic, and social science aspects.
Our goal is to have the model's outputs tailored to the multifaceted nature of climate change, thereby providing comprehensive insights essential for informed discussion and decision-making in the field.

To this end, we devise a three-step approach.
First, we utilize the ChatGPT \texttt{gpt-3.5-turbo} API to tag our retrieval database with labels corresponding to the natural science, economic, and social aspects.
Our preliminary experiments demonstrate that the quality of tags generated through few-shot prompts with \texttt{gpt-3.5-turbo} are satisfactory, and we do not further use \texttt{gpt-4}. The detailed prompt can be found in \Cref{sec:3_dimension_prompt}.
Second, during inference, we retrieve the most relevant documents for each dimension using the tags above.
These sets of documents are then fed separately to the model to generate three distinct completions.
The final step involves modifying the system prompt to instruct the model to focus on the specific dimension.
These special system prompts were already used during IFT training, for the examples where we had demonstrations focusing on these dimensions.

\section{Multilinguality}
\label{sec:multilinguality}

\begin{table}[ht!]
\begin{center}
\begin{tabular}{l|lll}
\toprule
Method & \multicolumn{3}{c}{Supported Languages} \\ \midrule
native in LLM & \multicolumn{3}{l}{English} \\ \midrule
cascaded MT & Arabic & Bengali & Chinese (simplified) \\
 & Dutch &  Finnish & French  \\
 & German & Greek & Hebrew \\
 & Indonesian & Japanese & Korean \\
 & Lithuanian & Pashto & Persian \\
 & Portuguese & Russian & Spanish \\
 & Thai & Turkish & Vietnamese \\ \bottomrule
\end{tabular}
\end{center}
\caption{\label{tab:supported_lang}Supported languages. English is native to the underlying LLM, while the support for other languages is achieved via a cascaded translation approach, i.e. \texttt{xx}$\rightarrow$\texttt{en} at the input and \texttt{en}$\rightarrow$\texttt{xx} at the output.}
\end{table}

To achieve our goal of making climate science accessible to a broader range of users, it is important that the model is not only accessible in English but is multilingual.
To support multilinguality in LLMs, there are two options. The first is to include multilingual data in the pre-training and IFT data. The other is to rely on Machine Translation (MT) to translate the user input into English, and to translate the generated text back into the user's language, i.e. a cascaded approach. We chose the latter because there is a lack of multilingual data in the climate domain and we wanted to maintain consistent quality in multiple languages. In \Cref{tab:supported_lang}, we list the supported languages.

To enable higher translation quality for climate text, we performed several domain adaptation experiments. Building on a generic base NMT model, we continued fine-tuning the model on parallel sentence pairs extracted from climate data only.
As the initial version of the model was presented in December 2023 at the 28th United Nations Climate Change conference in Dubai, we focussed our experiments on Arabic.

\subsection{In-Domain Data}

To extract parallel climate-related data from our large background collection of public and proprietary datasets, we explore two methods.

For the first method, we used around 2K climate terms and their human translations. We filtered the parallel data based on exact matches of these terms. We then fine-tuned our base model on these parallel sentences. We use Exact Match (EM) to denote this model. 

The second method is based on sentence embedding similarity. For this method, we take climate-related monolingual text, mainly a subset from our LLM pre-training data, as seed data. 
First, we use a weighted average over the word embeddings of a sentence to generate a fixed-size sentence embedding. To obtain a sentence pair embedding, we concatenate the source and target sentence embedding of each bilingual sentence pair. Afterwards, we employ k-Means clustering in the sentence pair embedding space. After obtaining a set of clusters, we use the in-domain seed data to determine which clusters should be used for training. This is done by selecting all clusters that contain a non-negligible portion of the in-domain data using a fixed threshold. For details, refer to AppTek's submission to the shared task of the IWSLT\footnote{International Workshop (Conference) on Spoken Language Translation.} evaluation~\cite{bahar-etal-2020-start}. The resulting parallel corpus is then used for fine-tuning the baseline translation model. We call this approach Embeddings Clustering~(EC).

In \Cref{tab:mt_data_stats}, data statistics related to the machine translation adaptation are shown. Initially, we took a subset of our training data, excluding those corpora that most likely will not benefit the conversation about climate. For example, we excluded subtitling data, transcriptions, and some others. The remaining data is the \emph{Filtered Base} data in \Cref{tab:mt_data_stats}. We filtered this data again to extract EM and EC data.

\begin{table}
\centering
\begin{tabular}{lrr}
\toprule
Data & Line count & Word count \\ \midrule
Filtered base & 88.4M & 1.9B \\
Exact Match & 2.5M & 63M \\
Embedding Clustering & 22.5M & 0.5B \\
IPCC Testing & 5.7K & 68K \\ \bottomrule \\[0.05em]
\end{tabular}\caption{\label{tab:mt_data_stats}Data statistics related to machine translation adaptation.}
\end{table}

\subsection{Training}

We used Transformer big architecture and parameters for the MT model \cite{NIPS2017_3f5ee243}. The fine-tuning process is done for a fixed number of steps for both data setups.
The training stopped after 15M parallel sentence pairs with a learning rate of $8.0\cdot 10^{-5}$.
This means that the EM data were seen multiple times, while the training did not go through all of the EC data.
For model adaptation usually the data size is much lower, and it does not make sense to train on a huge amount of data.

\subsection{MT Evaluation}

To construct a climate-domain test set, we used a translated IPCC report \cite{IPCC2014}. These reports have professional translations into multiple languages. We converted the PDF reports into text and aligned the sentences using vecalign \cite{thompson-koehn-2020-exploiting}.

Apart from evaluating our adapted MT models on this in-domain test set, we also computed automatic MT evaluation measures on a general-domain FLORES test set~\cite{nllb2022} and a test set extracted from OpenSubtitles\footnote{http://www.opensubtitles.org/}. Since climate-related questions and answers may have a broad range of styles and topics, we wanted to make sure that the adapted models did not over-fit to the style of the IPCC reports, but were still able to translate more general climate-related content well. 

\Cref{tab:mt_eval} presents the \textsc{Bleu} scores of the EM and EC models on the three evaluation sets. Here, we notice an improvement on the in-domain test data as expected, while degraded performance on the out-of-domain test set especially for the EM model can also be seen. This decrease in performance is not as large for the EC model. One probable explanation is that data filtering on exact matches is more strict than similarity-based filtering. From these results, we decided to go with the EC approach for the reverse translation directions, i.e. from English to Arabic.

\begin{table}
    \begin{center}
    \begin{tabular}{lrrr}
    \toprule
          & IPCC & FLORES & OpenSubtitles \\
    \midrule
         Base & 28.1 & 40.3 & 29.1\\
         EM &  29.5 & 39.7 & 26.9 \\
         EC & 28.8 & 40.0 & 28.1 \\
    \bottomrule
    \end{tabular}        
    \end{center}
    \caption{\textsc{Bleu} scores in \% of the baseline and adapted Arabic$\rightarrow$English models, using in-domain climate data or out-of-domain data as held-out evaluation sets.}
    \label{tab:mt_eval}
\end{table}

\subsection{Glossary Adaptation}

To make sure that climate-specific terms are almost always translated correctly, we use the climate term base that has formed the seed data for parallel data selection for the EM model also at inference time. We convert it to a glossary and use a glossary override method in which the desired glossary-based translation (a word or a phrase) is encoded together with the source term in the source sentence, both in training and at inference time. Our approach is similar to the glossary override method suggested by \cite{dinu-etal-2019-training}, with a difference that in training, randomly selected bilingual phrase pairs, determined via statistical word alignment, are inserted as artificial glossary entries. We also use special translation factors to mark the source and target glossary terms in the sentence, based on our factored NMT architecture~\cite{wilken2019novel}. With the successful adaptation of the MT models to the climate domain, the models are then used in a cascaded approach to enable multilinguality of the system, i.e. \texttt{xx} $\rightarrow$ \texttt{en} on the user query and \texttt{en} $\rightarrow$ \texttt{xx} on the LLM response.

\section{Automatic Evaluation}
\label{sec:automatic_evaluation}

Automatic evaluation of LLMs presents significant challenges that stem not only from the inherent complexities of natural language tasks but also from the difficulty in accurately capturing the multifaceted capabilities of LLMs with limited metrics. Despite these challenges, automatic evaluations, with their simplicity, interpretability, availability, and cost-effectiveness can serve as valuable proxies for assessing model performance.

In our evaluation process, we utilize both established tasks and custom tasks integrated into the LM Evaluation Harness \cite{eval-harness}, a widely adopted resource in large language model evaluations, as seen in platforms like the Open LLM\footnote{\url{https://huggingface.co/spaces/HuggingFaceH4/open_llm_leaderboard}}.
Our primary evaluation format involves text classification and multiple-choice questions (MCQs), structured as log probability ranking tasks. In the case of MCQs, the question and its corresponding choices are concatenated for model assessment. 
For a comprehensive overview of these tasks, please refer to \Cref{tab:eval_overview}. We publish all prompt templates and instructions on how to reproduce the evaluation\footnote{\url{https://github.com/eci-io/climategpt-evaluation}}.

\begin{table}
    \begin{center}
    \begin{tabular}{lllr}  \toprule
        Aspect &  Datasets & Test Set Size & Eval. Metric \\ \toprule
Climate-Specific& ClimaText & 1.6K & Accuracy \\
                        & ClimateStance & 0.3K & Accuracy\\ 
                        &  ClimateEng & 0.3K & Accuracy\\
          & ClimateFever & 1.5K & Accuracy \\
                          & CDP-QA  & 1.1K & Accuracy \\
                         & Pira 2.0 MCQ & 0.2K & Accuracy \\ 
                         & Exeter Misinformation & 2.9K & F1-Macro \\ \midrule
                General Domain  & PIQA & 1.8K & Accuracy\\
                        & WinoGrande & 1.2K & Accuracy \\
                        & MMLU & 14.0K & Accuracy \\ 
                        & HellaSwag & 10.0K & Normalized Accuracy \\
                        & OpenBookQA & 0.5K &  Normalized Accuracy \\
                        \bottomrule
    \end{tabular}
\end{center}
   \caption{Benchmarks in Climate and General Domains for downstream task evaluation and their corresponding metrics.}
    \label{tab:eval_overview}
\end{table}

\subsection{Climate-Specific Benchmarks}
\label{sec:climate_specific_benchmarks}

\paragraph{ClimaBench}~\cite{spokoyny2023answering} consists of a collection of diverse climate-related datasets designed to systematically evaluate model performance across a range of classification tasks. We employ the following 5 datasets from ClimaBench for evaluation: (i) \textbf{ClimaText}~\cite{leippold2020climatext}: This is a binary classification dataset containing sentences from the web, Wikipedia and the section of US public companies' 10-K reports that address climate-related risks. The task is to predict whether a given sentence is relevant to climate change or not. 
(ii) \textbf{ClimateStance}~\cite{vaid-etal-2022-towards} contains climate change-related tweets that were posted during the 2019 United Nations Framework Convention on Climate Change. The tweets have been manually categorized into three groups for the purpose of stance detection: those expressing support for climate change prevention, those opposing it, and those with an ambiguous stance.
(iii) \textbf{ClimateEng}~\cite{vaid-etal-2022-towards} is also a climate change related Twitter dataset, collected in the same manner as ClimateStance, for the task of fine-grained classification into topics such as: Disaster, Ocean/Water, Agriculture/forestry, Politics, General.
(iv) \textbf{ClimateFever}~\cite{leippold2020climatefever} is a fact-verification dataset of climate change-related claims. Consisting of 1,535 claims obtained from the Internet, each claim is paired with five pertinent evidence passages extracted from Wikipedia.   Each claim-evidence pair is labelled into one of three categories: Supports, Refutes, or Not Enough Info. We use this dataset in two ways for fact-verification: with evidence and without. In the first case (dubbed Fever-Evidence), the task is to detect whether certain evidence supports or refutes the claim or neither. The second task (dubbed Fever-Boolean) is to classify if a certain claim is true or false (without providing any evidence).  
(v) \textbf{CDP-QA}~\cite{spokoyny2023answering} is a dataset compiled from the questionnaires of the Carbon Disclosure Project, where cities, corporations, and states disclose their environmental information. The dataset presents pairs of questions and answers, and the objective is to predict whether a given answer is valid for the corresponding question. 
\paragraph{Pira 2.0 MCQ}~\cite{pirozelli2023benchmarks} is constructed using a compilation of scientific abstracts and United Nations reports focusing on the ocean, the Brazilian coast, and climate change. The objective involves choosing the correct answer from a set of five candidates in response to a given question, with or without supporting text. The candidate answers are carefully crafted to exhibit substantial lexical similarity with the supporting text and closely resemble the correct answer. This deliberate design adds an extra layer of complexity to the task, demanding a more profound comprehension of the question at hand.

\paragraph{Exeter Misinformation}~\cite{coan2021computer} dataset contains text from 33 influential climate contrarian blogs and climate change-related content from 20 conservative think tanks spanning the years 1998 to 2020. Annotation of the dataset was done manually using a thorough three-layer taxonomy of contrarian claims related to climate change, developed by the authors. We utilize this dataset specifically for the binary classification task of discerning whether a given text contains a contrarian claim on climate change or not. 

\subsection{General Domain Benchmarks}
Besides climate-specific benchmarks, we also conduct evaluations in the general domain. The goal is to maintain proficiency across general benchmarks while enhancing performance on climate-specific tests. The evaluation of models in the general domain focuses on two key aspects: commonsense reasoning and world knowledge. \\
The reasoning capabilities of models within the general domain are examined using four datasets: (i) \textbf{HellaSwag}~\cite{zellers2019hellaswag}: comprising multiple-choice questions derived from ActivityNet or wikiHow and challenging models to predict the next event in grounded situations; (ii) \textbf{PIQA}~\cite{Bisk2020}: containing binary-choice questions that require understanding of real-world object interactions in physical scenarios, (iii) \textbf{OpenBookQA} (O.B.QA)~\cite{OpenBookQA2018}: containing multiple-choice elementary-level science questions that evaluate the understanding of scientific facts and the ability to apply them to novel scenarios; (iv) \textbf{WinoGrande}~\cite{ai2:winogrande}: containing sentences in fill-in-the-blank format for the task of resolving ambiguous pronouns, given two options for completion. \\
The evaluation of models' world knowledge is carried out using the \textbf{MMLU} dataset~\cite{hendryckstest2021}, which consists of 57 subjects spanning STEM, humanities, social sciences, and beyond, with varying difficulty levels from elementary to advanced professional.

\subsection{Results}

\begin{table}
\begin{center}
\begin{tabular}{lrrr|r} \toprule
    Model             & ClimaBench & Pira 2.0 MCQ & Exeter Misinf. & Weight. Avg.  \\ \midrule
    Stability-3B & 71.4 & 48.7 & 52.6 & 62.8 \\
    Pythia-6.9B
    & 63.6 & 22.9 & 48.9 & 50.8 \\
    Falcon-7B & 62.9 & 19.8 & 39.9 & 48.3\\
    Mistral-7B & 73.1 & 80.0 & 63.7 & 73.7 \\
    Llama-2-7B & 68.5 & 51.1 & 59.4 & 62.6 \\
    Jais-13B & 66.9 & 26.4 & 54.2 & 54.4 \\  
    Jais-13B-Chat & 65.8 & 66.3 & 61.3 & 65.3 \\ \midrule
    Llama-2-Chat-7B & 67.8 & 72.0 & 64.3 & 68.5 \\
    Llama-2-Chat-13B & 68.6 & 79.3 & 68.6 & 71.4 \\
    Llama-2-Chat-70B & 72.7 & 88.8 & 72.5 & 77.0 \\ \midrule
    ClimateGPT-7B & 75.3 & 86.6 & 65.9 & 77.1 \\
    ClimateGPT-13B & 75.0 & 89.0 & 70.0 & 78.0 \\
    ClimateGPT-70B & 72.4 & 89.9 & 73.4 & 77.2 \\ \midrule
    ClimateGPT-FSC-7B & 59.3 & 17.2 & 45.1 & 46.2 \\
    ClimateGPT-FSG-7B & 53.1 & 17.4 & 41.5 & 42.1 \\
    \bottomrule
\end{tabular}
\end{center}
\caption{Results on the climate benchmarks.}
\label{tab:climate_automatic_results}
\end{table}

We perform an automatic evaluation on all of our final ClimateGPT models as well as a set of baselines and other publicly available foundation models.
As baselines, we consider the Llama-2 Chat models by Meta that were instruction fine-tuned on general domain data as well as further tuned using reinforcement learning from human feedback.

\Cref{tab:climate_automatic_results} shows the 5-shot results of all models on the set of climate-specific benchmarks.
To increase readability, we here just report the weighted average of ClimaBench results, the weights\footnote{Weights are assigned based on the nature of the task and its relevance to the practical application of the model.} and results for individual tasks are shown in \Cref{sec:full_auto_eval_results}. The first half of the table shows a comparison among foundation models: Stability-3B (\texttt{stabilityai/stablelm-3b-4e1t}), Pythia-6.9B (\texttt{EleutherAI/pythia-6.9b}), Falcon-7B (\texttt{tiiuae/falcon-7b}, Mistral-7B (\texttt{mistralai/Mistral-7B-v0.1}), Llama-2-7B (\texttt{meta-llama/Llama-2-7b-hf}) and Jais-13B (\texttt{core42/jais-13b}). We observe that Mistral-7B shows the best performance, followed by Stability-3B and LLama-2-7B. As mentioned in Section~\ref{sec:cpt}, due to the unavailability of Mistral-7B at the time of development, we continued with LLama-2-7B as our base model.

Next, we compare Llama-2 chat models with Llama-2-based ClimateGPT models and observe that all Llama-2-based ClimateGPT models outperform the corresponding Llama-2 Chat variant.
ClimateGPT-7B even outperforms the two times larger Llama-2 13B Chat model and performs on par with the 70B Chat model.
The two from-scratch models significantly underperform the Llama-2-based models.
While this was expected as Llama-2 was trained on significantly more data (2T compared to 0.3T tokens), we hoped that the potentially higher data quality of our corpus could counteract this.

We note that ClimateGPT-70B performs worse than expected on the climate-specific benchmarks and even worse than ClimateGPT-13B.
As discussed in \Cref{sec:cpt}, we did not have enough time at the end of the project to optimize hyper-parameters for the 70B models, so we assume the results can be significantly improved by additional optimization (e.g. lower learning rates).

Further, we observe that the FSC model outperforms the FSG model on climate-specific tasks.
While this gives an indicator that including domain-specific data already during pre-training could result in better results than the CPT of a general domain model, the difference is not large enough to justify training domain-specific models from scratch.
On the contrary, when considering the resource requirements of from-scratch training, this confirms the CPT approach used for the main ClimateGPT models.

\begin{table}
    \begin{center}
    \begin{tabular}{lrrrrr|r} \toprule
        Model             & PIQA & WinoGrande&HellaSwag & O.B.QA &MMLU & Avg.  \\ \midrule
        Stability-3B      & 79.4 & 66.3 & 76.1 & 40.8 & 44.6 & 61.5 \\
        Pythia 6.9B       & 76.6 & 61.2 & 65.5 & 36.2 & 25.9 & 53.1 \\
        Falcon-7B         & 80.3 & 67.3 & 78.1 & 43.6 & 27.1 & 59.3 \\
        Mistral-7B        & 81.8 & 73.9 & 83.4 & 48.0 & 63.5 & 70.1 \\
        Llama-2-7B & 79.0 & 69.1 & 79.0 & 45.2 & 47.0 & 63.9 \\
        Jais-13B & 76.5 & 68.4 & 73.1 & 38.6 & 35.0 & 58.3 \\
        Jais-13B-Chat     & 76.5& 68.4& 73.1&38.6&68.4& 63.1   \\ \midrule
        Llama-2-7B-Chat   & 79.0& 69.1& 79.0& 45.2& 69.1& 62.9 \\
        Llama-2-13B-Chat  & 79.9& 72.3&82.3 &48.2&72.3& 66.4   \\
        Llama-2-70B-Chat  & 83.9 & 78.0& 87.0&52.2&78.0& 68.6  \\ \midrule
        ClimateGPT-7B     &79.8 &70.3& 78.4&47.6& 68.6& 65.1   \\
        ClimateGPT-13B    & 80.7&73.4& 82.0&51.8&73.1& 68.8    \\
        ClimateGPT-70B    & 83.6&79.4&85.8&53.0 &66.6& 73.7    \\ \midrule
        ClimateGPT-FSC-7B & 72.9& 53.4& 58.9&36.0&23.0& 48.8   \\
        ClimateGPT-FSG-7B & 72.5& 54.5&58.7&38.6&25.1& 49.9    \\
        \bottomrule
    \end{tabular}
    \end{center}
    \caption{Results on the general benchmarks.}
    \label{tab:general_automatic_results}
\end{table}

\Cref{tab:general_automatic_results} shows the results of the general domain benchmarks for our baselines and the ClimateGPT models. We report 10-shot results on HellaSwag and 5-shot on all other benchmarks.
We used these benchmarks to verify that our models do not over-fit on the climate domain and still perform on par with comparable models on non-climate tasks.
Comparing the main ClimateGPT models to the Llama-2 Chat models, we observe that we not only do not degrade but even outperform the baseline models.
Most of our general benchmarks are still science focused so we assume that the additional climate data also benefits these benchmarks.

\subsection{Cascaded Machine Translation}

\begin{table}
    \begin{center}
        \begin{tabular}{lrr}
            \toprule
             & \multicolumn{2}{c}{EXAMS (Acc [\%])} \\
            Model & Arabic & MT Ar-En \\ \midrule
            Llama-2-13B-Chat & 25.0 & 38.0 \\
            Llama-2-70B-Chat & 28.7 & \textit{38.6} \\
            Jais-13B-Chat & \best{40.9} & 36.0 \\
            \bottomrule
        \end{tabular}
    \end{center}
    \caption{Arabic EXAMS evaluation results of the English-only Llama-2 Chat models and compared with the bilingual model Jais, taking the exam directly in \emph{Arabic} and on an \emph{Ar-En} machine translation of the dataset.}
    \label{tab:eval_cascaded}
\end{table}

To evaluate the cascaded machine translation approach we evaluate on the Arabic subset of the EXAMS dataset \cite{hardalov-etal-2020-exams}.
EXAMS is a multiple choice question answering collected from high school examinations.
The Arabic subset covers questions from biology, physics, science, social science and Islamic studies.

For this evaluation we compare Llama-2 Chat which was primarily trained on English text to the bilingual Arabic-English Jais Chat model (\texttt{core42/jais-13b-chat}) \cite{sengupta2023jais}.
All models are first evaluated directly on the original \emph{Arabic} version of the datasets.
Second, we translate the questions and answer options using our adapted \emph{Ar-En} MT system (see \cref{sec:multilinguality}) from Arabic to English.
The results are shown in \Cref{tab:eval_cascaded}.
We first observe that when evaluating directly on Arabic, as expected, the Llama-2 Chat models significantly underperform Jais Chat.
But, when making use of cascaded machine translation, the Llama-2 Chat models recover most of the performance gap compared to Jais.
At the same time with Jais, we observe a degradation in the results when using the cascaded MT approach from 40.9\% in Arabic to 36.0\% in English.
An explanation for that is, as shown in \cref{tab:general_automatic_results}, Jais performs worse than Llama-2 Chat on English benchmarks (i.e. 65.3 for Jais compared to 71.4 for Llama-2-Chat-13B).
Another explanation could be errors or potential ambiguities in the automatic translation.

Overall, this shows that cascaded MT is a promising direction to scale LLMs to a large number of languages.
While truly native LLMs for each specific language seem to perform better, training LLMs for each specific language or a truly multilingual LLM would require a large amount of resources.

\section{Human Evaluation}
\label{sec:human_evaluation}

The value of automatic evaluations is their very strict definition as well as their repeatability which allows precise comparisons of systems and models. Automatic evaluations need a set of clearly defined references and a metric that allows one to measure the difference between the system output and the reference. For tasks like speech recognition the reference is clear (the sequence of words that have been spoken) and counting substitutions, insertions and deletions is fairly easy. For machine translation (respectively summarisation), because, given a source sentence, several equivalent translations having the same meaning are possible, flexible metrics have been developed like \textsc{Bleu} \cite{papineni-etal-2002-bleu} (respectively \textsc{Rouge} \cite{lin-2004-rouge} for summarisation). For question answering, (e.g. answering questions on climate change) the variability in the formulation of responses is much higher than that for translation or summarisation of text where the original text serves as base. Domain knowledge, factuality and the bringing of arguments in a certain order (reasoning) are key to the quality of an answer. While the tasks used for our automatic evaluation do cover in some respects climate change and climate science knowledge they do not give us indications of how well an answer is formulated, essentially they do not give us insights on how well the arguments (facts) of the answer are brought together, how good the reasoning supported by the arguments do lead to an easy to understand conclusion. Multiple-choice questions (used for automatic evaluations as they allow to limit the number of valid outputs) are good at verifying the understanding of atomic knowledge. ClimateGPT being targeted at answering complex questions has to generate the dots between the acquired atomic knowledge, bringing in the reasoning needed to make the answer self-explainable. It is probably necessary to master the knowledge behind a domain like climate science to answer complex questions on climate, but it is not enough. A good answer depends on the ability to reason on this knowledge and on the formulation clarity of this reasoning. As of today, quality evaluation of answers to questions is best done with humans having a comprehensive overview of the field and who can judge whether a response is adequate, if it covers all relevant aspects to the question and if the reasoning supported by the information provided is well formulated/expressed.

For our human evaluation, a set of 7 climate change post-docs, PhD students and master students has been asked to provide feedback on the output of 3 different models by ranking them against each other, Also they were asked to tell us if some claims in the generated outputs have been hallucinated or not.
For the ranking, human evaluators had positive and negative points to distribute according to the following principles:
\begin{itemize}
\item Evaluate the quality of each answer by ranking them within each other and also qualifying the goodness of each answer.
\item Positive numbers are good, negative numbers are bad, the zero is neutral.
\item Refer to the sheet \enquote{Quality Dimensions} for your evaluation.
\item If answers differ only in their syntactical form, please consider them equal. 
\item The answers in each column have been randomly taken from one of the 3 system outputs so that you shall not be tempted to find a pattern per system or that you do not develop a preference for a system.
\item Edit columns B C and D from the \enquote{Ranking} sheet according to the following principles:
\item You will have the following numbers at your disposition:
\begin{itemize}
 \item                2  /  1  /  0  /  -1  /  -2
 \end{itemize}
\item The negative numbers are bad grades. 
\begin{itemize}
\item 2 is best
\item 0 is average
\item -2 is the worst
\end{itemize}
\item As we want to rank the answers, try to avoid giving the same rank to 2 system outputs.
\item If all are of the same quality, all get the same grade, 
between 2 if all are very good and -2 if all very bad.
\item If 2 are similarly good and 1 is bad, the 2 good ones get a positive number (e.g. 1)  and the 3rd a negative number (e.g. -1)
\item If 2 answers are good and one is better, the better one gets a 2 and the less better one a 1
\end{itemize}

The \enquote{Quality dimensions} referred to in the above guidelines are those from \cite{bulian2023assessing}.

We noticed that human evaluators tended to try to find patterns for each system. To increase the neutrality of their judgment, we decided not to name each answer by the model name and to randomly order the answers of each system so that the annotators were not tempted to try to guess from which system each answer comes.

\subsection{Results}

\begin{table}
    \begin{center}
    \begin{tabular}{lrr} \toprule
        Model             & Average Rank & \# Hallucinations \\ \midrule
        ClimateGPT-70B      & 1.0 & 2\\
        ClimateGPT-7B      &0.6 & 4\\
        ClimateGPT-FSC-7B      &0.2 & 5\\
        \bottomrule
    \end{tabular}
    \end{center}
    \caption{Human evaluation comparing the answers of 50 questions from 3 different systems. An average rank around 0 means the system has been evaluated half of the time as good and half of the time as not good, independently of its rank.}
    \label{tab:human_evaluation_results}
\end{table}

For the human evaluations, we asked the evaluators to compare and rank three versions of our ClimateGPT models: the from-scratch model ClimateGPT-FSC-7B, the CPT models ClimateGPT-7B and ClimateGPT-70B.
This evaluation shows us that CPT models compare positively against the from-scratch model (1.0 points vs. 0.2), and the 70B CPT model performs better than the 7B CPT model (see table \ref{tab:human_evaluation_results} first block).
While the ranking between from-scratch and CPT models correlates well with the automatic evaluation, it is not the case when comparing the 7B and 70B within the CPT model family.
We need to investigate this further.
Another outcome of this evaluation is the observation that the lower the rank of a model, the higher the number of hallucinations.

\section{Responsible AI}

The pursuit of responsible AI systems is a critical aspect as important as, if not more than, the model performance itself. In this work, we aim to follow closely the standard approaches in the field. Of course, as an active and evolving field of study, the definition and scope of ``responsible AI" continue to develop in tandem with the advancement of more sophisticated AI systems.

\subsection{Content Moderation}
\label{sec:content_moderation}

Our perspective on responsible AI encompasses two fundamental aspects: maximizing benefit and minimizing harm. This reflects an inherent trade-off between a model's usefulness and its safety. In the realm of LLMs, for instance, a system that refrains from answering any question minimizes risk but offers limited utility, whereas a system that responds to all queries increases usefulness but may be prone to misuse like generation of misinformation. A pertinent example is content moderation. Simple methods like keyword block lists, as used in the Jais model \cite{sengupta2023jais}, can be effective: a safe refusal message is triggered by a regex check against a predefined word list. However, such approaches risk having too many false positives, for example, the keyword 'sex', though potentially problematic, can be a part of legitimate biological discussions. This illustrates how surface-level safety measures might inadvertently constrain a model's utility.

A more elegant solution is to fine-tune the model on data that gracefully answers unintended contents.
In our case, we adopt the Do-Not-Answer dataset \cite{wang2023donotanswer}, and manually check many responses from the baseline Llama-2-Chat 70B model \cite{touvron2023llama}. The model responses are often satisfactory, i.e. not only refusals but also include helpful explanations and suggestions. Encouraged by this, we decide to augment the Do-Not-Answer dataset with these model completions and include it in our IFT dataset (see \Cref{sec:ift_general}).
Additionally, this automated approach to the curation of content moderation examples spares human annotators from the stress of handling toxic data just to replicate Meta's existing efforts, further aligning with our responsible AI principles.
While this approach effectively helps to reduce undesired outputs and reduces the potential for misuse of the model, we want to note that these fine-tuning approaches can be easily circumvented if an attacker has access to the model \cite{yang2023shadow,zhan2023removing}.

\subsection{Transparency}

Transparency is a cornerstone of responsible AI, fostering reproducibility, facilitating communication, and revealing potential issues. The recent introduction of the Foundation Model Transparency Index (FMTI) \cite{bommasani2023foundation} offers a framework to assess the transparency of foundation models. 

Although we agree that the specific questions and their weightings in FMTI may be subject to debate \cite{fmti-critique}, it represents a significant step towards standardizing disclosure practices in LLM research. In our work, we nonetheless reference FMTI to self-assess and achieve an FMTI score of 69 and also self-assessed using the revised methodology and achieved a score of 62.\footnote{Detailed results: \url{https://github.com/eci-io/climategpt-fmti}}

These self-evaluations underscore our commitment to sustainability and the ongoing discourse towards transparent model development.

\subsection{Environmental Impact}

The environmental footprint is a critical consideration in responsible AI, especially for projects in the climate domain.
Recognizing the substantial economic and computational resources required for training LLMs, we prioritized the use of sustainable energy sources.
In collaboration with MLFoundry, we accessed a high-performance computing cluster powered exclusively by hydroenergy.
Although securing high-end GPU computing resources, especially those powered by green energy, is challenging, our decision to partner with a provider committed to clean energy sources reflects our dedication to minimizing the environmental impact of our work.

Using a framework developed by \citet{hershcovich-etal-2022-towards}, we have published a sustainability scorecard that details the energy usage and emissions associated with training experiments, as well as the final models.
You can find the scorecard in the appendix in \Cref{tab:sustainability_scorecard}.
This data is currently being evaluated by Filecoin Green to establish a Green Score\footnote{\url{https://www.greenscores.xyz/}} and will be published once finalized.

\section{Conclusion}

This paper has introduced ClimateGPT, a domain-specific large language model (LLM) that gives access to interdisciplinary research information on climate change. ClimateGPT is therefore the first family of LLMs to generate not only one answer but four different answers, three answers each along a different perspective (science/economic/social which we also call dimension) plus a fourth answer summarising the answer of the three perspectives provided to the user. We have compared five different ClimateGPT versions. The first two models are from scratch (FS) foundational models trained on our own 300B tokens corpus, both with a 7B parameters transformer architecture similar to that of Llama-2. The training corpus of the first FS model contains 4B climate change data (Climate-FSC-7B), the other do not have these 4B tokens (Climate-FSG-7B) which allows us to compare the value of climate change data within a foundational model. The next three models are based on a Llama-2 foundational model (7B, 13B and 70B, all pre-trained on 2T tokens).  These  models as well as the ClimateGPT-FSG-7B are fine-tuned on 4B climate tokens (Continued Pre-Training, CPT). We did not fine-tune the ClimateGPT-FSC-7B model as it had these 4B tokens in the from-scratch training.
All models have been further trained on a manually defined Instruction Fine Tuning (IFT) climate-specific prompt/completion pair corpus that has been produced by experts (climate consultants and climate scientists) and by non-experts. We have been benchmarking our models on two different sets of tasks, one set specific to climate (ClimaBench, Pira and Exeter), and the other one on standard non-climate tasks. We have shown that, while adapting our models to climate change we do not lose performance on general tasks (MMLU, HellaSwag, PIQA and WinoGrande) and that our Llama-2-based ClimateGPT-7B outperforms Llama-2-Chat-13B (77.1\% resp. 71.4\%) on climate tasks with two times fewer parameters and is on par with the Llama-2-Chat-70B results (77.1\% resp. 77.0\%) with 10 times less parameters. 

The quality of the IFT data plays an important role. An interesting question that we did not have time to address is whether a general science IFT dataset would also have contributed and by how much.

In this paper, we show also the value of cascaded machine translation as opposed to using a general one-system fits-all-languages approach. The comparison made on the Arabic subset of EXAMS between the mono-lingual system Llama-2-13B-Chat with that of Jais-13B-Chat (which has been also trained on Arabic data) shows that Machine Translation (translating the EXAMS from Arabic to English, so translating the query and the answer into English) allows Llama2-13b-Chat to improve from 25.0 to 38.0 very near the performance of Jais-13b-Chat (40.9) which was evaluated directly in Arabic.
We did not fine-tune our MT for this specific task.

Another important aspect of our work is related to the sustainability of domain-specific models: fine-tuning our ClimateGPT-7B has been done with a tiny fraction (so small that it is apparent to a rounding error) of the CO2 production needed to produce the complete Llama-2-Chat-70B model. Further, at inference time, our system answers questions producing 12 times less CO2 (needing 12 times less energy) than Llama2-70B would do, for the same result. 

Finally, our human evaluations show some correlation with the set of automated tasks used for benchmarking.

\section{Limitations}
\label{sec:limitations}
Like any LLM, ClimateGPT is subject to hallucinations.
Retrieving relevant documents for grounding before calling ClimateGPT can help control hallucinations.
While subjectively our proposed RAG approach seems to reduce hallucinations, we have not yet performed a systematic evaluation.

This interdisciplinary project involving multiple partners is challenging in itself and inevitably has many limitations. Here, we acknowledge the drawbacks by following standard practices in the NLP community and hope to inspire future work.

Firstly, while Reinforcement Learning from Human Feedback (RLHF) is an effective method for enhancing model performance, we do not employ it due to time and resource constraints. We notice the model gives decent performance without RLHF, and thus we focus our efforts elsewhere.

Secondly, LLMs are shown to exhibit strong native multilinguality and can perform the machine translation task very well.
However, we apply a cascaded approach to make use of domain fine-tuned existing MT systems at AppTek because fully training a multilingual LLM would inevitably be much more costly.

We also acknowledge the limitation of not using a domain-specific tokenizer which could have improved the model's representation of climate-related vocabulary as discussed in \Cref{sec:from_scratch}.

For retrieval, we relied on a standard bi-encoder model and did not investigate domain adaptation or more advanced retrieval techniques like reranking or verification of the relevance of retrieved documents.

Finally, another limitation is in evaluation. Automatic evaluation is limited in reliability and what they can evaluate and our human evaluation is limited in scale, completeness, and breadth.
Therefore, a lot of design decisions still need to be validated with systematic evaluations.

\section*{Acknowledgments}
The model development and evaluation was completed as independent research in advance of the COP28 Conference from August to December 2023 through a grant provided by ADQ, TAQA, Masdar, Etihad Rail, ADNEC Group, and Hedera.

We would like to thank The Club of Rome for their partnership and unwavering support, specifically Sandrine Dixson-Declève, Paul Shrivastava, Mmampele Rampele, Peter Blom, Carlos Alvarez Pereira, Wouter van Dieren, Per Espen Stoknes, Jorgen Randers, Gunter Pauli; Nature Finance, specifically Simon Zadek; Goals House specifically Matthew Freud, Arlo Brady, Anna Biles; Info.nl specifically Jann de Waal, Dominik Vrbic, Anandita Punj, Jorrit Tinholt,  Paul Domen; The Internet Archive, Brewster Kahle, Mark Graham, Wendy Hanamura, Jamie Joyce, as well as the support from Babiche Veenendaal-Westerbrink.

Special thanks to David Lobell for participating in our expert interviews which gave invaluable insights in his work and are the foundation of our IFT data.
Next, we want to thank Acheampong Baafi-Adomako, Hamidreza Mosaffa, Pan Hao, Qinghua Yu, Ralf Liebermann, Thomas Kreuzwig and Yurong Gao for creating the expert IFT dataset as well as for having participated in our human evaluations.
We further want to thank Eugen Beck, Nico Daheim, Nils Hilgers and Ege Beysel for helpful discussions on this work.

We thank ML Foundry for the opportunity to train ClimateGPT using renewable energy, giving us early access to their H100 machines and for all the support across different time zones during the main training phase.

In addition, we'd like to acknowledge Faisal Al Hammadi and the entire team at Further Ventures, who helped coordinate the sponsorship funding and supported our vision to bring the model to COP28.

We would also like to thank the engineering and cryptography team at EQTY Lab that worked on the integrity of the AI lifecycle:
Benedict Lau,
Yurko Jaremko,
Paul Dowman,
Cameron Fyfe,
Tyler Brink,
Mauve Signweaver,
Tucker McCoy and
Ziv Weissman. And a special thanks to Dan Boneh.

Further, the team at Gladeye for designing and creating our website: 
Tarver Graham,
Conrad Blight,
Nathan Walker,
Antony Zouch,
Kate Forsythe,
Alastair Gray,
Michael Cannon,
Giuliana Aliotti,
Daniel Bonham,
Joris Rotteveel and
William Hamlin.

And all the people who worked on the responsible AI pipeline and pilots:
Judie Muhrez,
Alex Feerst,
Chris DiBona,
Monica Granados and the entire Creative Commons team,
Marc Johson (Filecoin Green),
Dr. Regina Stanback Stroud (RSSC),
Anton Blewett,
Travis Coan (Exeter),
John Cook (University of Melbourne),
Nathan Schneider,
Joshua Tan,
Connor Spelliscy,
Scott Moore and
Khalifa University, Students: Benhur Tekeste, Divora Yemane, Maryam Alblooshi, Maryam Alraeesi and Noof Alhammadi.

Finally, this work would not have been possible without various contributions from the open-source community.
We want to highlight the Open Assistant project for crowd-sourcing a high-quality multi-turn IFT dataset, Meta for sharing Llama-2, the EPFL LLM Team for their work on Megatron-LLM and finally Databricks for sharing Dolly.
 
\bibliography{anthology,custom}
\bibliographystyle{acl_natbib}
\newpage
\appendix

\section{Appendix}

\subsection{Model Card}

\Cref{tab:model_card} presents a model card \cite{mitchellModelCards} that summarizes details of the models.

{\renewcommand{\arraystretch}{1.35}
\begin{table}[htbp]
\begin{center}
\scalebox{0.82}{
\begin{tabular}{ p{3cm}|p{15cm}  }
 \toprule
 \multicolumn{2}{ c }{\textbf{Model Details}} \\
 \midrule
 \textit{Model Developers} & AppTek, EQTYLab, Erasmus AI \\
 \hline
 \textit{Variations} & ClimateGPT comes in a range of parameter sizes: 7B, 13B, and 70B. Additionally, there are two 7B model variants trained from scratch. \\
 \hline
 \textit{Input} &  Models input text only. \\
 \hline
 \textit{Output} & Models generate text only. \\
 \hline
\textit{Model Architecture} &ClimateGPT is an auto-regressive language model that uses an optimized transformer architecture. After pre-training, instruction fine-tuning (IFT) is used to align the models to the expected output format. \\
\hline
 \textit{Model Dates} & ClimateGPT was trained between September 2023 and November 2023. \\
 \hline 
 \textit{Status} & This is a static model trained on an offline dataset and intended to dynamically include new knowledge via RAG. \\
 \hline
 \textit{License} & ClimateGPT Community License \\
 \hline
 \textit{Where to send comments} & Feedback can be given by creating a discussion thread on the model's Huggingface page (\url{https://huggingface.co/eci-io/}). \\
 \midrule
 \multicolumn{2}{ c }{\textbf{Intended Use}} \\
 \midrule
 \textit{Intended Use Cases} & ClimateGPT is intended to be directly used as a question-answering model that is specialized in the climate domain. It is built to provide useful feedback for decision-makers, scientists and journalists involved in climate discussions. \\
 \hline
 \textit{Out-of-Scope Uses} & Use in any manner that violates applicable laws or regulations. \\
  \midrule
  \multicolumn{2}{ c }{\textbf{Hardware and Software} (\Cref{sec:pre_training})} \\
\midrule
\textit{Training Factors} &
   We used a fork of the Megatron-LLM Repository by the EPFL LLM Team (\url{http://github.com/epfLLM/Megatron-LLM}). A cluster provided by MLFoundry was used for pre-training, instruction fine-tuning and evaluation.\\
  \hline
\textit{Carbon Footprint} &  Pre-training utilized a cumulative 31,059 GPU hours of computation on hardware of type H100 SXM (including CPU TDP of 775W). The cluster for training and evaluation was powered using 100\% hydropower (24g CO$_2$eq/KWh \cite{ipcc2014hydro}) which resulted in the emission of 577.7kg CO$_2$eq.\\
 \midrule
  \multicolumn{2}{ c }{\textbf{Training Data} (\Cref{sec:pretraining_data,sec:ift}) } \\
\midrule
 \textit{Overview} & ClimateGPT was continuously pre-trained on a dataset of 4.2B climate-specific tokens. The from-scratch models were trained on 300B tokens. All models were instruction fine-tuned on a dataset consisting of public IFT data as well as IFT data collected in cooperation with climate experts during the project. \\
 \hline
 \textit{Data Freshness} & The pretraining data contains documents up to October 2023. \\
 \midrule
 \multicolumn{2}{ c }{\textbf{Evaluation Results}} \\
  \midrule
  \multicolumn{2}{ p{18cm} }{
  See automatic evaluation in \Cref{sec:automatic_evaluation} and human evaluation in \Cref{sec:human_evaluation}
 } \\
 \midrule
\multicolumn{2}{ c }{\textbf{Ethical Considerations and 
Limitations} (\Cref{sec:limitations})} \\

 \midrule

\multicolumn{2}{p{18cm}}{

Despite the efforts from the development team to eliminate them, as with every other chat-capable LLM, this model may generate biased, offensive or inaccurate responses. Testing done to date has been mostly in English and no extensive red-teaming was conducted. Therefore, for all downstream applications users should be made aware of these limitations and should be incentivised to double check model outputs.}
\\
 \bottomrule
 
\end{tabular}}
\end{center}
\caption{Model card for ClimateGPT.}
\label{tab:model_card}
\end{table}}

\newpage

\subsection{Sustainability Scorecard}

\begin{table}[htbp]
\begin{center}
\begin{tabular}{lr}
\toprule
Model Publicly Available &  Yes \\
Time to train final models & 31,059 GPU Hours \\
Location for computations for final models & United States (WA) \\
Energy mix at location for final models & 24 gCO2eq/kWh \\
Power of GPU and CPU for final models & 0.775 kW \\
CO2eq for final models & 577.70 kgCO2eq \\
Time for all experiments & 1,535 GPU Hours (Canada, ON) \\
&2,150 GPU Hours (United States, CA)\\
Power of GPU and CPU for experiments & 0.55 kW \\
Location for computations for experiments & Canada, ON \& United States, CA  \\
Energy mix at location for experiments & 134 gCO2eq/kWh \& 186 gCO2eq/kWh \\
CO2eq for all experiments & 113.13 kgCO2eq \& 219.95 kgCO2eq \\
Average CO2eq for inference per sample & 24.5 mgCO2eq \\
\bottomrule
\end{tabular}
\end{center}
\caption{Sustainability scorecard for ClimateGPT.}
\label{tab:sustainability_scorecard}
\end{table}

\newpage

\subsection{Curated Climate-Specific Pre-Training Data Details}
\label{sec:appendix_climate_data_details}

The following section gives additional details on the high-quality and manually curated climate-specific datasets that are part of our pre-training data.

\paragraph{Extreme Weather Events}
A corpora built out of the the most recent decade (2023-2013) of extreme weather news reports, in excess of on average 1M articles per year (slightly less in the earlier years and more in the latter).
The intent of the corpora is to build a collection of human-centered effects of climate change, and how extreme weather events impact human activity systems. 

The articles were categorized using a custom-built classifier from Erasmus.AI which from its daily planetary scale web crawl organized the articles into 19 categories (Drought, Sandstorm, Extreme Heat Wave, Forest Fire, Wildfire, etc.) and one not-relevant class. 
Low certainty scoring articles were eliminated from training corpora.  

Candidate articles were translated into English from 19 languages.
Extreme weather events in certain areas such as West Africa, Latin America, and parts of China do not have a great deal of reporting in English.
For example in 2023 in Columbia, 78\% of the articles were reported in Spanish.
In Peru, 97\% of the articles were non-English and were collected from 128 unique websites and associated with 442 Peruvian cities.

The candidate articles were geographically assigned using named entity recognition and a proprietary framework from Erasmus.AI to ensure higher accuracy in assigning events to specific locations as well as to check the events against each other as it would be highly unlikely to have single reports of extreme weather events.

The events were matched to human timelines, and reporting about expected future weather events was eliminated, as much of climate reporting relates to future events that may or may not happen.

The overall goal of the corpora was to train the model family on how, at a human activity systems level, the changing climate connects with geographic knowledge as well as deeper knowledge on the causal effects of climate events (e.g. Snowstorm leads to electricity outages, stay at home orders, supply disruptions; Drought in the Horn of Africa leads to increased civil conflict, etc.).

\paragraph{Technical Game-Changing Breakthroughs}
For Europe's largest technology company Erasmus.AI in partnership with the Digital Thinking Network researched and identified 153 game-changing breakthroughs in Energy, Climate Change, Food Security, Health, etc..
The process encompassed 500+ pages of technical documentation, and 153 themes set up in a proprietary interface NewsConsole run by Erasmus.AI.
The interface enables graph-based curation of large bodies of articles through multiple views (narrative analysis, temporal, etc. views).
A theme might present breakthroughs in super-capacitors, desalination technologies, multiple approaches to batteries, novel bacteria that convert sunlight directly into animal feed, saltwater-based agriculture, etc..
Each of these themes presents a forward-looking approach to addressing climate challenges with technology.
This includes, for example, not just experience curves in Solar PV and Battery technology, but the viewpoint that these experience curves will continue to make Solar PV (and wind) the cheapest forms of energy in most locations or breakthroughs in animal feed.
A selection of the top-ranked few thousand articles per theme was used, where ranking was a combination of human feedback, automated systems, and curation on rich visual interfaces.

\paragraph{Sustainable Development Goals}
For the Club of Rome as pre-work into the Earth4All process, Erasmus.AI prepared a breakdown of the 17 Sustainable Development Goals (SDGs) of the United Nations into sub-goals and set up a framework of themes on the platform described above that tracks these subgoals of the SDGs.
The intent here is to provide a more holistic human-scale view of climate change, and its effects, where action on carbon reduction using cutting edge Solar PV and Battery technologies is constrained by for example poverty.
Action on climate change is not just simply the rational allocation of resources to enable the best long-term returns for a healthy planet and humanity.
The climate change action discussion is deeply political with countries in the Global South making the case that they will be bearing the brunt of the climate liabilities in which historically polluting countries in the Global North have accrued the benefits.
Capturing these nuances in terms of the 17 SDGs and a holistic outcomes-driven discussion seemed prudent in data selection to ensure the model has some degree of the human development challenges inherent in the climate change discussion.

\paragraph{Climate Change News}
Next to the Extreme Weather Corpora, Erasmus.AI searched and curated corpora based on a set of climate semantic concepts.
The semantic concepts were built and curated initially from themes inside Erasmus.AI's NewsConsole which displays machine- and human-curated visualisations of narrative analysis of vast amounts of articles.
Once the set of these themes was deemed of sufficient quality through human inspection, these concepts were used for larger-scale searches through the Erasmus.AI corpora.

\paragraph{Climate Change Specific Corpora}
International development organizations, treaty organizations, and the broad NGO community (World Bank, OECD, IPPC, UN, EU, TCFD, US Gov, Nation State Governments, NASA, ESA, WRI, IREA, WEF, Nature Finance, etc.) together publish significant well-researched work on climate change and its impacts on financial systems, countries, ecologies, etc.
Erasmus.AI built a collection of these reports from a combination of existing collections and performed a set of custom crawls.

\paragraph{Climate Academic Research}
A set of academic publications of open access and open web academic articles were collected on climate change.
It was decided to limit this corpus to open access and open web full articles (and not just abstracts) to ensure that the model represented logical arguments, inherent in full academic publications not just conclusions taken from abstracts.

\newpage

\subsection{AppTek Non-Expert IFT Data Details}
\label{sec:annotator_demographic}

\begin{table}[htbp]
    \begin{center}
    \begin{tabular}{lr}
    \toprule
    \textbf{Age} & \textbf{Count} \\
    \midrule
    18-29 & 72 \\
    30-49 & 25 \\
    50-69 & 2 \\
    \bottomrule
\end{tabular}
\end{center}
    \caption{Age distribution of non-expert IFT data annotators.}
    \label{tab:apptek_annotator_age}
\end{table}

\begin{table}[htbp]
    \begin{center}
\begin{tabular}{lrr}
    \toprule
    \textbf{Country} & \textbf{Num. Annotators} & \textbf{Num. prompts} \\
    \midrule
    India & 51 & 4970 \\
    USA & 35 & 1901 \\
    Germany & 6 & 560 \\
    Mexico & 3 & 3090 \\
    CAN & 3 & 107 \\
    Costa Rica & 1 & 200 \\
    \bottomrule
\end{tabular}
\end{center}
    \caption{Geographic distribution of non-expert IFT data annotators.}
    \label{tab:apptek_annotator_origin}
\end{table}

\begin{table}[htbp]
    \begin{center}
\begin{tabular}{lrr}
    \toprule
    \textbf{Topic} & \textbf{\%} \\
    \midrule
    Climate & 9.2 \\
    \multicolumn{2}{l}{\textbf{From Experts Interviews}} \\
    Central Bank Policies & 2.3 \\
    Extreme Weather & 2.4 \\
    Geo-engineering & 2.3 \\
    Industrial Systems & 2.0 \\
    Natural Systems and Services & 2.3 \\
    Reducing Carbon Emissions & 2.6 \\
    Regenerative Agriculture & 0.2 \\
    \multicolumn{2}{l}{\textbf{Other Topics}} \\
    Agriculture & 2.3 \\
    Animals & 3.9 \\
    Culture & 4.0 \\
    Ecosystems & 4.1 \\
    Energy & 4.1 \\
    Environment & 5.5 \\
    Health & 3.7 \\
    History & 3.9 \\
    Legal & 4.1 \\
    News & 3.6 \\
    Politics & 4.0 \\
    Technology & 4.1 \\
    Travel & 3.9 \\
    Weather & 24.0 \\
    \bottomrule
\end{tabular}
\end{center}
    \caption{Distribution of topics provided to non-expert IFT data annotators during the data collection process.}
    \label{tab:topic_distribution}
\end{table}

\newpage

\subsection{Retrieval Augmentation Example}

\begin{table}[ht!]
\centering\normalsize
\begin{tabular}{c|c|p{0.68\textwidth}}
\hline
 & Rank & Text \\ \hline
user query & - & What policies should India implement in Kolkata to mitigate flooding? \\ \hline
\multirow{5}{*}{retrieved chunk} & 1 & land use and land cover of ekw between 2000 and 2019 when engaging with officials and engineers from the kolkata municipal corporation and the i \& wd, it becomes evident that wastewater regulation is influenced by various pressures that must be considered in urban environmental planning, management, and governance. officials utilize the bantala lock gate to lower the wastewater level in the main canal, especially during monsoons, to accelerate sewage and stormwater flow and mitigate the risk of urban flooding. the dense informal settlements in the deltaic city of kolkata heighten the threat of waterlogging, and \\ \cline{2-3} 
 & 2 & heighten the threat of waterlogging, and the deterioration of both natural and human - made blue infrastructure amplifies the city's vulnerability to floods. officials are cautious about attracting media attention and criticism if flood risks are not managed effectively. however, opening the bantala lock gate hinders wastewater supply to the inlet canals and, consequently, the bheris. this puts kolkata municipal corporation and the i \& wd in a dilemma, torn between addressing the needs of low - lying urban residents and fishers during the monsoons. the media portrayal suggesting a deliberate \\ \cline{2-3} 
 & 3 & the monsoons. the media portrayal suggesting a deliberate jeopardizing of fishing livelihoods to facilitate real estate development is likely an exaggeration. decisions made by other government agencies and departments emphasize that there is no systematic effort to convert the wetlands into built - up areas. \\ \cline{2-3} 
 & 4 & these floods is likely to increase as the climate changes, particularly due to storm surges, sea level rise and more intense precipitation. " future proofing " kolkata against climate change, population growth and economic development is an immense challenge, particularly considering the scale of poverty and informality in the city. iii. methods this paper evaluates the implications of " business - as - usual " modes of development for kolkata ’ s energy use, energy bills and greenhouse gas emissions in the period to 2025. it also evaluates a wide range of energy efficiency, renewable energy and other mit \\ \cline{2-3} 
 & 5 & kolkata and other indian cities are experimenting with more inclusive forms of urban planning and policymaking. kolkata has a tropical climate, with monthly mean temperatures varying from 19 to 30°c. most rainfall occurs during the monsoon season between june and september. the city frequently experiences flooding during this time due to the inadequate drainage and sewer networks, which do not serve the city ’ s whole population. where this infrastructure exists, it is often a century old and lacks the capacity to meet the current population ’ s needs. the frequency and severity of these floods is likely to increase as the climate changes \\ \hline
\end{tabular}\caption{Examples retrieval results.}
\end{table}

\newpage

\subsection{System Prompts}
\label{sec:appendix_system_prompts}

\begin{table}[ht]
  \begin{center}
    \begin{tabular}{p{40mm}p{90mm}} \toprule
        \textbf{Subset} & \textbf{System Prompt} \\ \midrule
        Senior Expert Interviews & You're ClimateGPT a large language model synthesizing inter-disciplinary research on climate change. Always answer as helpfully and professional as possible, while being safe. Avoid colloquial language. Your answers should not include any harmful, unethical, racist, sexist, toxic, dangerous, or illegal content. Please ensure that your responses are socially unbiased. \\
        \midrule
        Grounded Expert Demonstrations & You're an expert in climate science. Always answer as helpfully and professional as possible, while being safe. \\ 
        \midrule
        Grounded Non-Expert Demonstrations & You're a helpful assistant supporting users with their questions on climate change.\textbackslash n Cite the documents provided in the context. \\
        \midrule
        StackExchange & You're an AI assistant generating answers to questions on the website stackexchange on the topic \{source\}. \\
        \midrule
        AppTek General & You're a helpful and harmless AI assistant. \\
        \midrule
        OASST-1 & You're Open Assistant, an AI language model, developed by Laion AI together with an open source community and trained using crowdsourced data. \\
        \midrule
        Dolly & You're an AI language model trained on data generated by employees of databricks. \\
        \midrule
        Llama-2 Safety & You're a helpful assistant supporting users with their questions on climate change. \\
        \midrule
        FLAN & You're a multi-task model solving a variety of NLP tasks. Give short responses only and follow the format of the user query. \\
        \midrule
        CoT & You're a multi-task model solving a variety of NLP tasks. Give short responses only and follow the format of the user query.  \\  
        \bottomrule
    \end{tabular}
     \end{center}
    \caption{System prompts used for IFT training for each of the different subsets.}
   \label{tab:appendix_system_prompts}
\end{table}

\newpage

\subsection{Prompts used in Climate-Specific Automatic Evaluation Tasks}
\begin{table}[ht]
  \begin{center}
    \begin{tabular}{p{35mm}p{65mm}} \toprule
        \textbf{Task} & \textbf{Prompt} \\ \midrule
        ClimaText & Given the following statement, is it relevant to climate change or not:\textbackslash n\{\}\textbackslash nAnswer: \\ \midrule
        ClimaStance & Is the following statement in-favor, against, or ambiguous about climate change prevention:\textbackslash n\{\}\textbackslash nAnswer:\\ \midrule
        ClimateEng & Given the five categories: 'general', 'politics', 'ocean/water', 'agriculture/forestry', 'disaster', assign the following statement to one of the categories: '\{\}'. \textbackslash nAnswer:\\ \midrule
        CDP-QA & Given a question and an answer, examine if the answer addresses the question.\textbackslash nQuestion: \{\}\textbackslash nAnswer: \{\}\textbackslash n\textbackslash nOutput:\\ \midrule
        Fever-Boolean & Is the following statement on climate correct or misinformation:\textbackslash n\{\}\textbackslash nAnswer: \\ \midrule
        Fever-Evidence & Given the following documents:\textbackslash n\{\}\textbackslash n\textbackslash nIs the following claim:\textbackslash n\{\}\textbackslash nSupported or Refuted?\textbackslash n \\ \midrule
        Pira 2.0 MCQ (no context) & Answer the following question with the correct alternative. GIVE ONLY THE CORRECT LETTER. \textbackslash nQuestion: \{\}. A: \{\}. B: \{\}. C: \{\}. D: \{\}. E: \{\}\textbackslash nANSWER:\\ \midrule
        Pira 2.0 MCQ (with context) & Based on the following context: \textbackslash n\textbackslash n \{\}. \textbackslash n Answer the following question with the correct alternative.  GIVE ONLY THE CORRECT LETTER \textbackslash n Question: \{\{question\}\} \textbackslash n\textbackslash n A: \{\}. B: \{\}. C: \{\}. D: \{\}. E: \{\}\textbackslash nANSWER: \\ \midrule
        Exeter Misinformation & This is a climate-misinformation classification task. Your task is that of telling whether the given text presents a contrarian claim regarding climate change. Your reply should be: 1: contains a contrarian claim; 0: does not contain a contrarian claim. Your reply should contain only the corresponding number and nothing else (i.e., 0 or 1).\textbackslash nTEXT: \{\} ANSWER: \\ \bottomrule
    \end{tabular}
     \end{center}
    \caption{Prompts used in each of the climate-specific automatic evaluation tasks.}
   \label{tab:setup}
\end{table}

\newpage

\subsection{Prompt for Retrieval Database Tagging}\label{sec:3_dimension_prompt}

In Figure \ref{tagging_prompt}, we present the prompt we used with the OpenAI \texttt{gpt-3.5-turbo} text completion API to tag text chunks from our retrieval database.
The text chunks in the few-shot example sections in the prompt all come from the IPCC Climate Change 2014 Mitigation of Climate Change report \cite{ipcc2014climate}.
Our initial runs with the prompt proved to be satisfactory, but not perfect.
For example, despite adding ``Please generate comma-separated, plain-text tags, e.g. natural,social (no need to add space after the comma separators and do NOT repeat your tags).", the model sometimes did not follow this specific instruction exactly.
Nonetheless, such imperfections are easily fixable with post-processing scripts, and we stuck with the prompt.

\begin{figure}[!t]
\begin{center}
\scalebox{0.85}{\noindent\makebox[\textwidth]{%
  \fbox{%
    \parbox{\textwidth}{%
\small Your task is to tag a chunk of text with labels from ("natural", "economic", "social"), depending on which science discipline the text is closest to. A chunk of text can have multiple tags. Please generate comma-separated, plain-text tags, e.g. natural,social (no need to add space after the comma separators and do NOT repeat your tags). Below, you will be presented with some examples. Each example is formatted as:

\# text

<text to be tagged>

\# tags

<natural|economic|social>...

Examples will follow "=== example ===" and the text to be tagged will follow "=== to be tagged ===".\\[0.5em]

=== example ===

\# text

5.2.3.5 Sulphur dioxide and aerosols
Uncertainties in SO2 and carbonaceous aerosol (BC and OC) emissions have been estimated by Smith et  al. (2011) and Bond et  al. (2004, 2007). Sulphur dioxide emissions uncertainty at the global level is relatively low because uncertainties in fuel sulphur content are not well correlated between regions. Uncertainty at the regional level ranges up to 35\%. Uncertainties in carbonaceous aerosol emissions, in contrast, are high at both regional and global scales due to fundamental uncertainty in emission factors. Carbonaceous aerosol emissions are highly state-dependent, with emissions factors that can vary by over an order of magnitude depending on combustion conditions and emission controls. A recent assessment indicated that BC emissions may be substantially underestimated (Bond et al., 2013), supporting the literature estimates of high uncertainty for these emissions.

\# tags

natural\\[0.5em]

=== example ===

\# text

The energy intensity of the industry sector could be directly reduced by about 25\% compared to the current level through the wide-scale upgrading, replacement and deployment of best available technologies, particularly in countries where these are not in use and in non-energy intensive industries (high agreement, robust evidence). Additional energy intensity reductions of about 20\% may potentially be realized through innovation (limited evidence, medium agreement). Barriers to implementing energy efficiency relate largely to initial investment costs and lack of information. Information programmes are a prevalent approach for promoting energy efficiency, followed by economic instruments, regulatory approaches and voluntary actions. [10.7, 10.9, 10.11]

\# tags

economic\\[0.5em]

=== example ===

\# text

Reduction of subsidies to fossil energy can achieve significant emission reductions at negative social cost (very high confidence).  Although political economy barriers are substantial, many countries have reformed their tax and budget systems to reduce fuel subsidies that actually accrue to the relatively wealthy, and utilized lump-sum cash transfers or other mechanisms that are more targeted to the poor. [15.5.3]

\# tags

social\\[0.5em]

=== example ===

\# text

No single factor explains variations in per-capita emissions across cities, and there are significant differences in per capita GHG emissions between cities within a single country (robust evidence, high agreement). Urban GHG emissions are influenced by a variety of physical, economic and social factors, development levels, and urbanization histories specific to each city. Key influences on urban GHG emissions include income, population dynamics, urban form, locational factors, economic structure, and market failures. Per capita final energy use and CO2 emissions in cities of Annex I countries tend to be lower than national averages, in cities of non-Annex I countries they tend to be higher. [12.3]

\# tags

economic,social\\[0.5em]

=== to be tagged ===

\# text

\textbf{\textit{<Text Chunk To Be Tagged>}}

\# tags
    }%
  }%
}}
\end{center}
\caption{\label{tagging_prompt} Prompt used for retrieval database tagging with the OpenAI \texttt{gpt-3.5-turbo} text completion API.}
\end{figure}

\newpage

\subsection{Full Automatic Evaluation Results}
\label{sec:full_auto_eval_results}
Table~\ref{tab:full_auto_eval_results} shows the performance of the models on the individual climate-specific tasks grouped under ClimaBench and Pira benchmarks. 

\begin{table}[h!]
    \scriptsize
    \centering
    \begin{tabular}{m{10em}m{1cm}m{1cm}m{1.2cm}m{1.2cm}m{1.2cm}m{1.2cm}m{1.1cm}m{1.2cm}} \toprule
       Models  & CDP-QA & Clima-Text & Climate-Eng & Climate-Stance & Fever-Boolean & Fever-Evidence & Pira-MCQ  (no ctx) & Pira-MCQ (with ctx) \\ \midrule
        {Weights} & 1.0 & 0.5 & 0.5 & 0.5 & 1.0 & 1.0 & 1.0 & 1.0 \\ \midrule
        Stability-3B & 74.7 & 57.1 & 61.1 & 80.6 & 72.2 &  74.9 & 40.5 & 56.8 \\
    Pythia-6.9B & 67.8 & 52.8 & 36.6 & 78.0 & 63.1 & 71.6 & 21.6 & 24.2 \\
    Falcon-7B & 78.7 & 57.7 & 48.2 & 75.8 & 50.5 & 63.0 & 21.1 & 18.5 \\
    Mistral-7B & 79.3 & 66.7 & 65.4 & 78.0 & 71.7 & 72.9 & 67.0 & 93.0 \\
    Llama-2-7B & 73.1 & 55.8 & 61.1 & 72.7 & 66.3 & 74.0 & 45.8 & 56.4 \\
    Jais-13B & 67.1 & 62.9 & 60.8 & 56.9 & 70.6 & 73.0 & 19.8 & 33.0 \\
    Jais-13B-Chat & 71.3 & 72.8 & 35.8 & 40.0 & 70.5 & 80.0 & 58.1 & 74.4 \\ \midrule
    Llama-2-Chat-7B & 77.4 & 72.1 & 60.8 & 70.1 & 62.0 & 64.2 & 63.0 & 81.1 \\
    Llama-2-Chat-13B & 72.2 & 74.7 & 52.7 & 62.3 & 69.7 & 72.0 & 68.3 & 90.3 \\
    Llama-2-Chat-70B & 77.0 & 78.1 & 59.4 & 69.9 & 70.8 & 75.7 & 83.3 & 94.3 \\ \midrule
    ClimateGPT-7B & 81.2 & 70.5 & 65.1 & 59.4 & 73.5 & 81.0 & 81.1 & 93.0 \\
    ClimateGPT-13B & 83.0 & 76.3 & 68.5 & 56.6 & 77.6 & 76.1 & 82.4 & 95.6 \\
    ClimateGPT-70B & 83.2 & 78.3 & 68.7 & 50.1 & 69.9 & 74.3 & 85.5 & 94.3 \\ \midrule
    ClimateGPT-FSC-7B & 43.7 & 46.5 & 53.0 & 77.7 & 63.5 & 70.9 & 18.1 & 16.3 \\
    ClimateGPT-FSG-7B & 35.1 & 50.0 & 45.4 & 77.7 & 45.2 & 72.3 & 20.3 & 14.5 \\ \bottomrule
    \end{tabular}
    \caption{Five-shot performance on climate-specific automatic evaluation tasks. Task-specific weights are used to compute the weighted-average score in Table~\ref{tab:climate_automatic_results}.}
    \label{tab:full_auto_eval_results}
\end{table}

\newpage

\subsection{MT Glossary Examples}

In \Cref{tab:example_mt_glossary}, we share a few examples of Glossary entries that are were used during the MT inference.

\begin{table}[ht!]
\begin{center}
\begin{tabular}{lr}
\toprule
Original & Adjustment/Correction (if applicable) \\ \midrule
Adaptation Research & \<أبحاث التكيف> \\
Paris Agreement & \<اتفاقية باريس> \\
Obligate Species & \<أجناس شبه محددة الموطن> \\
Water Stress & \<إجهاد المياه> \\
Carbon Capture and Sequestration & \<احتجاز الكربون وامتصاصه> \\
Carbon capture and storage & \<احتجاز الكربون وتخزينه> \\
Bond event & \<أحداث بوند> \\
Earth's energy imbalance & \<اختلال توازن الطاقة> \\
Eustatic Sea-Level Rise & \<ارتفاع مستوى سطح البحر> \\
Sea Level Rise & \<ارتفاع مستوى سطح البحر> \\ \bottomrule \\[0.01em]
\end{tabular}
\end{center}
\caption{Example climate-related glossary used during machine translation inference.}
\label{tab:example_mt_glossary}
\end{table}

\end{document}